\documentclass[10pt,twocolumn,letterpaper]{article}

\usepackage[pagenumbers]{iccv} 
\usepackage{bm}
\usepackage{multirow}
\usepackage{makecell}
\usepackage[table]{xcolor}
\usepackage{pifont}
\usepackage{tcolorbox}
\newcommand{\ra}[1]{\renewcommand{\arraystretch}{#1}}

\definecolor{iccvblue}{rgb}{0.21,0.49,0.74}
\usepackage[pagebackref,breaklinks,colorlinks,allcolors=iccvblue]{hyperref}

\def\paperID{1708} 
\def\confName{ICCV}
\def\confYear{2025}

\title{CC-Diff: Enhancing Contextual Coherence in Remote Sensing Image Synthesis}

\author{
    \begin{tabular}[t]{c}
    Mu Zhang \and 
    Yunfan Liu\thanks{Corresponding Author} \and
    Yue Liu \and 
    Yuzhong Zhao \and
    Qixiang Ye
    \end{tabular}
    \\
    University of Chinese Academy of Sciences, Beijing, China \\
    \tt\small{\{zhangmu23, liuyue171, zhaoyuzhong20\}@mails.ucas.ac.cn}, \tt\small{\{liuyunfan, qxye\}@ucas.ac.cn}
}

\begin{document}
\maketitle
\begin{abstract}

%
%

Existing image synthesis methods for natural scenes focus primarily on foreground control, often reducing the background to simplistic textures.
Consequently, these approaches tend to overlook the intrinsic correlation between foreground and background, which may lead to incoherent and unrealistic synthesis results in remote sensing (RS) scenarios.
In this paper, we introduce CC-Diff, a \underline{\textbf{Diff}}usion Model-based approach for RS image generation with enhanced \underline{\textbf{C}}ontext \underline{\textbf{C}}oherence.
Specifically, we propose a novel Dual Re-sampler for feature extraction, with a built-in `Context Bridge' to explicitly capture the intricate interdependency between foreground and background.
Moreover, we reinforce their connection by employing a foreground-aware attention mechanism during the generation of background features, thereby enhancing the plausibility of the synthesized context.
%
%
%
Extensive experiments show that CC-Diff outperforms state-of-the-art methods across critical quality metrics, excelling in the RS domain and effectively generalizing to natural images.
Remarkably, CC-Diff also shows high trainability, boosting detection accuracy by 1.83 mAP on DOTA and 2.25 mAP on the COCO benchmark.
\end{abstract}

%
%

\begin{figure}[t]
\centering
\includegraphics[width=0.95\linewidth]{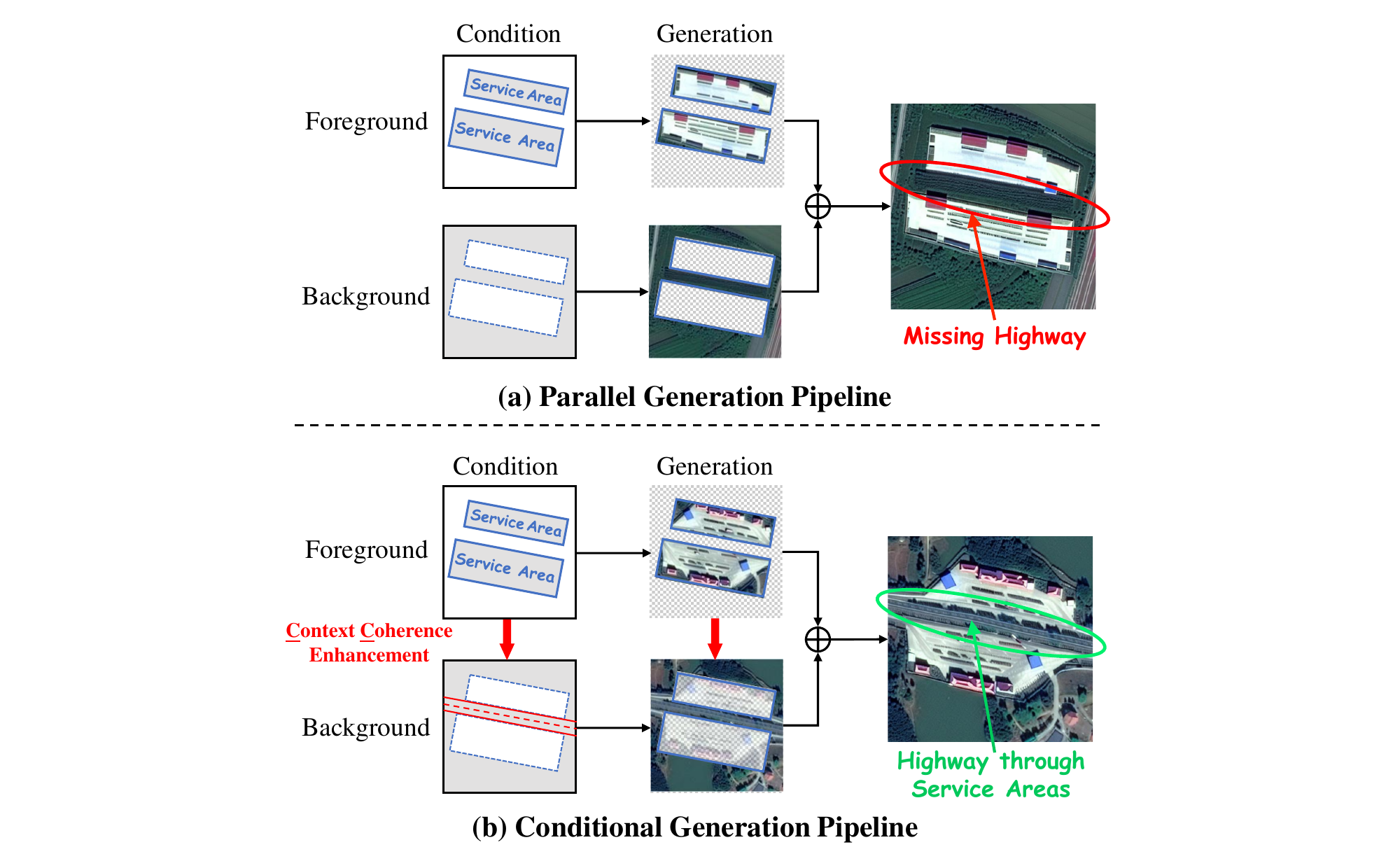}
\captionof{figure}{Comparison of (a) parallel and (b) conditional generation pipeline. The conditioning mechanism (denoted by red arrows) enhances the \textbf{contextual conherence} of generation results.}
\label{fig:sample_RS_synthesis}
\end{figure}

\section{Introduction}
\label{sec:intro}

Recent studies have highlighted the remarkable potential of synthetic imagery in boosting visual perception tasks~\cite{singh2024synthetic,wang2024generated,zhao2024generating,aydemir2025data}. 
Motivated by these promising outcomes, the Geoscience community has devoted increasing attention to controllable remote sensing (RS) image synthesis~\cite{xu2023txt2img,samar2024diffusionsat,tang2024crsdiff,sebaq2024rsdiff,yuan2023RSFSG}, aiming to improve accuracy in a variety of analytical tasks~\cite{wang2022rvsa,li2023lsknet,yu2024std,liu2023seeing,wang2024samrs,blumenstiel2024mess}.

While many existing efforts~\cite{xu2023txt2img,samar2024diffusionsat,sebaq2024rsdiff} rely on textual prompts to encode image semantics, these prompts often fail to capture essential spatial cues (\textit{e.g.}, position and orientation) of foreground objects. 
To address this gap, some researchers incorporate dense guidance (\textit{e.g.}, semantic maps) to enhance controllability~\cite{tang2024crsdiff,yuan2023RSFSG}, which inevitably raises annotation requirements and restricts both flexibility and diversity in the generated outputs.

\begin{figure}[t]
   \centering
    \includegraphics[width=0.95\linewidth]{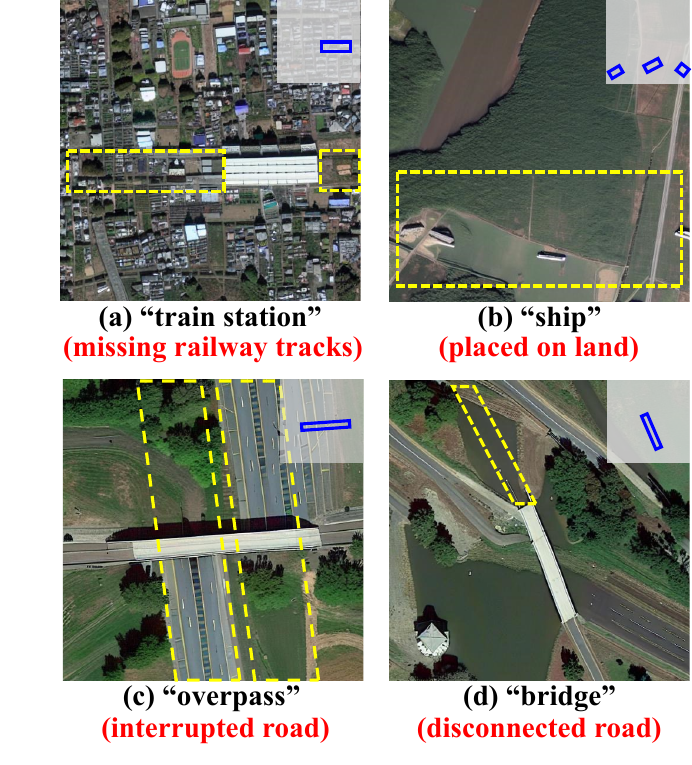}
    \caption{Illustration of contextual incoherencies in RS images synthesized by~\cite{zhou2024migc}. Layouts are shown in the top-right corner, object classes are labeled below with quotation marks, and incoherencies are highlighted with dashed yellow boxes.
    }
   \label{fig:Incoherency_Example}
\end{figure}

Motivated by recent advances in spatially controllable image generation, the Layout-to-Image (L2I) technique offers a promising solution to the aforementioned challenge.
However, most existing L2I methods~\cite{zheng2023layoutdiffusion,li2023gligen,zhou2024migc,zhou2024migc++,xie2023boxdiff} adopt a parallel generation pipeline (Figure~\ref{fig:sample_RS_synthesis} (a)), which primarily focuses on aligning the attributes of synthesized foreground instances (\textit{e.g.}, texture, color, position) with the given prompt, often overlooking broader contextual coherence with the background.
This oversight may be less critical in many natural image scenarios as the background usually acts as a simple backdrop for object placement, it becomes far more consequential in RS imagery, where foreground objects are closely interlinked with their environment. 
Hence, ensuring contextual coherence is vital for producing semantically consistent results (see Figure~\ref{fig:Incoherency_Example}).


To address this issue, we introduce CC-Diff, a \underline{\textbf{C}}ontext-\underline{\textbf{C}}oherent \underline{\textbf{Diff}}usion Model for RS image generation.
An in-depth analysis of existing L2I methods reveals that semantic inconsistencies primarily stem from separate, non-interacting modules handling foreground and background synthesis.
While CC-Diff maintains a multi-branch design, it differs from existing approaches by establishing a cross-module conditioning mechanism that links the two synthesis pipelines.
This mechanism enhances foreground awareness throughout both background feature extraction and rendering, as shown in Figure~\ref{fig:sample_RS_synthesis}(b).
Comprehensive experiments on RS datasets demonstrate that CC-Diff yields realistic, contextually coherent results with a high degree of controllability. 
In addition, CC-Diff exhibits strong trainability: it boosts detection accuracy by 1.83 mAP on the challenging DOTA dataset~\cite{xia2018dota} and by 2.25 mAP on the more generalized COCO benchmark~\cite{lin2014coco}, underscoring its robust generalizability.
%
%

Our contributions are summarized as follows:
\begin{itemize}
    \item We introduce CC-Diff, an L2I framework originally conceived to address the incoherence in RS imagery synthesis, yet readily applicable to broader image domains.

    \item We propose a conditional pipeline that explicitly captures the interdependence between foreground instances and their backgrounds, utilizing distinct learnable queries to integrate both background texture and the semantic context of the foreground.

    \item Extensive experiments show that CC-Diff not only generates realistic, semantically consistent images across both RS and natural domains, but also serves as a highly effective augmentation strategy for object detection tasks.
\end{itemize}

\section{Related Work}
\label{sec:related_work}

\noindent \textbf{Controllable Image Generation.}
%
%
%
Text-to-Image (T2I) and Layout-to-Image (L2I) are the two main categories of controllable image generation methods.
T2I approaches aim to synthesize images reflecting the semantics of textual descriptions. 
Early work in this field was dominated by Generative Adversarial Networks~\cite{reed2016generative,xu2018attngan,zhang2021cross}, until DALL-E~\cite{ramesh2021zero} demonstrated the potential of autoregressive frameworks. 
Subsequent research~\cite{ding2021cogview,yu2022parti,chang2023muse} further improved fidelity and scalability. 
Meanwhile, another line of work employs Diffusion Models~\cite{sohl2015deep,ho2020denoising} guided by text prompts to achieve realistic T2I synthesis~\cite{ramesh2022hierarchical,saharia2022photorealistic,Nichol2022glide,rombach2022high}.
In contrast, L2I methods generate images from instance layouts to ensure spatial precision. 
With the rise of Diffusion Models, L2I research~\cite{xie2023boxdiff, wang2024instancediffusion, zheng2023layoutdiffusion, li2023gligen, yang2023reco} increasingly focuses on integrating layout conditions. 
Recently, Multi-Instance Generation (MIG)~\cite{zhou2024migc, zhou2024migc++, chen2024geodiffusion, wu2024ifadapter} has gained traction by separating the generation of foreground elements from background content, thereby enhancing their attributes.

\noindent \textbf{Remote Sensing Image Synthesis.}
Although T2I for natural images has made significant progress, T2I in the RS domain remains in its infancy.
Early attempts include Txt2Img-MHN~\cite{xu2023txt2img}, which employs hierarchical prototype learning to bridge the semantic gap between text prompts and RS imagery, and DiffusionSat~\cite{samar2024diffusionsat}, which leverages LDM~\cite{rombach2022ldm} alongside textual and numeric metadata.
RSDiff~\cite{sebaq2024rsdiff} further enhances image quality through super-resolution.
While these methods produce visually plausible RS images, they struggle to precisely control attributes such as the spatial layout of foreground instances.
The contemporary study AeroGen~\cite{tang2024aerogen} introduces a layout-controllable diffusion model supporting rotated bounding boxes, and related efforts~\cite{yuan2023RSFSG,Miguel2023mapsat,sastry2024geosynth,tang2024crsdiff} add additional guiding information. 
However, none explicitly address maintaining coherence between foreground and background.
%

%
%
%
%
\section{Preliminaries}
\label{sec:preliminaries}

\subsection{Latent Diffusion Model}

The Latent Difffusion Model (LDM)~\cite{rombach2022high} enhances the computational efficiency of vanilla Diffusion Models~\cite{ho2020denoising, Nichol2022glide,ramesh2022hierarchical} by performing denoising in the latent space, and improving controllability through the integration of a cross-attention mechanism.
After obtaining the latent representation $\bm{z}\in Z$ of the image $\bm{x}$, LDM seeks to learn a denoising autoencoder $\epsilon_{\boldsymbol{\theta}}$, which progressively generating less noisy data $\bm{z_{T-1}}, \bm{z_{T-2}}, \dots, \bm{z_0}$ from an initial sampled noise $\bm{z_T}$.
In practice, $\epsilon_{\boldsymbol{\theta}}$ is trained to predict the step-wise noise $\boldsymbol{\epsilon}$ of the forward process, with the objective expressed as
\begin{equation}\label{eq:LDM}
\min_{\boldsymbol{\theta}} \mathcal{L}_{LDM} = \mathbb{E}_{\bm{z}, \boldsymbol{\epsilon}\sim\mathcal{N}(\bm{0},\bm{I}), t}[\|\boldsymbol{\epsilon}-\epsilon_{\boldsymbol{\theta}}(\bm{z}_t, \bm{c}, t)\|_2^2]
\end{equation}
where $t$ is a time step sampled from interval ${1,...,T}$ and $\bm{c}$ represents the prompt embedding.

\subsection{Cross-Attention Mechanism}

In LDM, the semantics of generated images are guided by conditional prompts through a cross-attention (CA) mechanism~\cite{rombach2022high}.
Given a latent image feature $\mathbf{f}$ and a condition $\mathbf{c}$, they are first projected to obtain the query, key, and value representations: $\mathbf{Q} = \mathbf{W_Q} \cdot \mathbf{f}$, $\mathbf{K} = \mathbf{W_K} \cdot \mathbf{c}$, and $\mathbf{V} = \mathbf{W_V} \cdot \mathbf{c}$, where $\mathbf{W_Q}$, $\mathbf{W_K}$, and $\mathbf{W_V}$ are learnable projection matrices.
The CA between $\mathbf{f}$ and $\mathbf{c}$ is then computed by 
\begin{equation}\label{eq:cross_attention}
\text{CA}(\mathbf{f}, \mathbf{c}) = \text{Softmax}\left(\frac{\mathbf{Q}(\mathbf{f})\mathbf{K}(\mathbf{c})^\top}{\sqrt{d}}\right)\mathbf{V}(\mathbf{c}),
\end{equation}
All subsequent cross-attention operations in this paper follow the formulation in Eq.~\ref{eq:cross_attention}.

\subsection{Problem Definition}

let $\mathcal{L}=\{\mathbf{c_i}, \mathbf{b_i}\}_{i=1}^N$ represent the layout containing $N$ foreground objects, where the $i$-th object is assigned a class label $\mathbf{c_i}$ and an oriented bounding box $\mathbf{b_i}=[x_i,y_i,w_i,h_i,{\theta}_i]$.
Specifically, $(x_i,y_i)$ denotes the coordinates of the top-left corner, $w_i$ and $h_i$ are the width and height, and ${\theta}_i$ is the orientation angle.
Moreover, an auxiliary textual description $\mathcal{P}$ is introduced to capture the image’s global semantics.
In this setting, CC-Diff learns a mapping $\mathcal{G} : (\mathcal{L}, \mathcal{P}, \mathbf{z}) \rightarrow \bm{I}$, where $\bm{I}$ represents the generated image semantically aligned with $\mathcal{L}$ and $\mathcal{P}$, and $\mathbf{z}$ denotes the Gaussian noise.
For details of the construction of $\mathcal{L}$ and $\mathcal{P}$, please refer to Sec. \ref{sec:def_angle} and Sec. \ref{sec:text_prompt} of the Appendix. 
\begin{figure}[t]
  \centering
   \includegraphics[width=1.0\linewidth]{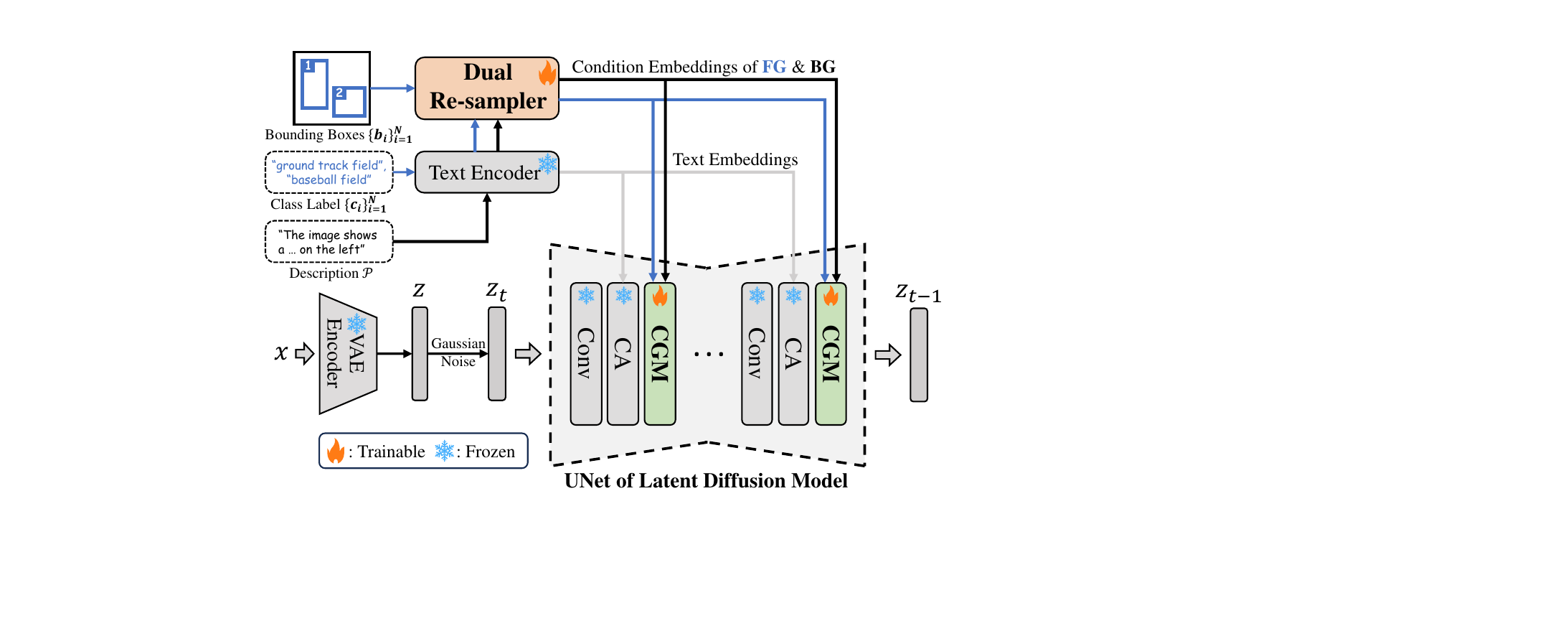}
   \caption{The framework of CC-Diff. The \textbf{Dual Re-sampler} extracts condition embeddings from user inputs (bounding boxes, class labels, and descriptions), guiding the \textbf{Conditional Generation Module (CGM)} to produce contextually coherent outputs.
   }
   \label{fig:framework}
\end{figure}

\begin{figure*}[th]
    \centering
    \includegraphics[width=0.93\linewidth]{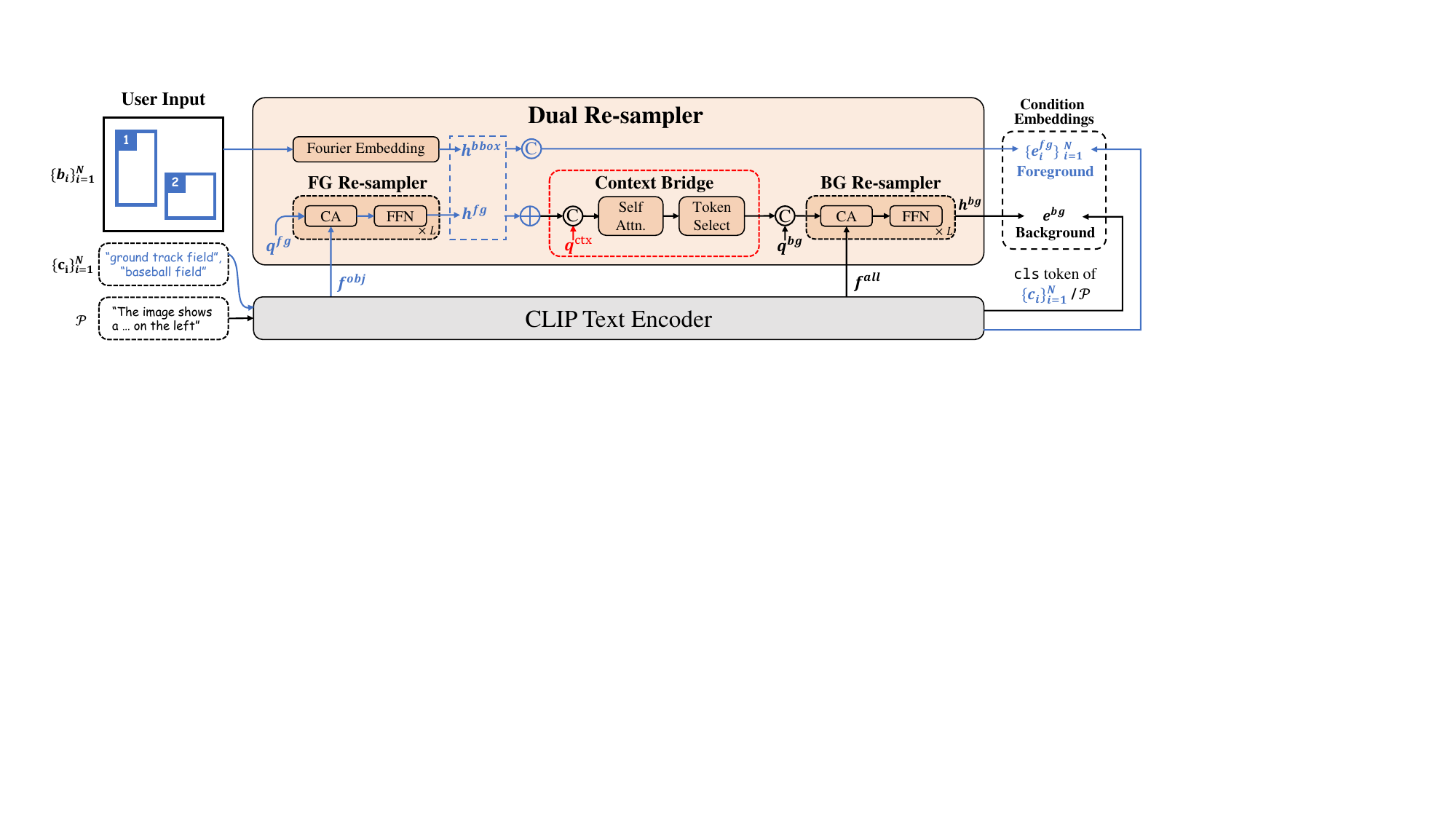}
    \caption{The architecture of Dual Re-sampler. The context query $\mathbf{q^{ctx}}$ obtains contextual semantics of FG objects from the FG Re-sampler, then incorporates them into BG feature extraction within the BG Re-sampler, thereby establishing the FG-BG association.
    }
    \label{fig:dual_resampler}
\end{figure*}

\begin{figure}[t]
    \centering
    \includegraphics[width=0.95\linewidth]{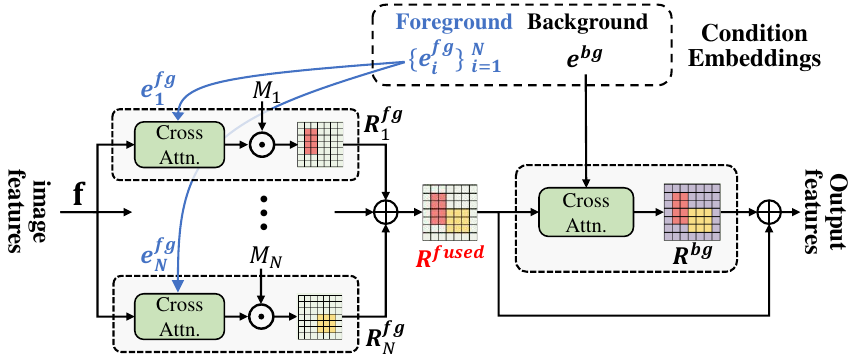}
    \caption{The architecture of Conditional Generation Module (CGM). The BG feature is rendered using the fused FG representation $\mathbf{R^{fused}}$, ensuring FG-awareness throughout generation.}
    \label{fig:cgm}
\end{figure}

\section{Method}
\label{sec:method}

\subsection{Overview}

As illustrated in Figure~\ref{fig:framework}, CC-Diff employs a Dual Re-sampler to encode the semantic information of foreground (FG) and background (BG), along with their relationships, into condition embeddings. 
These embeddings are then passed to the \textbf{Conditional Generation Modules (CGMs)} within the UNet of LDM to guide the generation with enhanced contextual coherence. 
Detailed introductions of these two modules are provided as follows.

\subsection{Dual Re-sampler for FG-BG Association}
\label{sec:dual_resampler}

As its name suggests, the Dual Re-sampler comprises two specialized re-sampler modules for extracting FG and BG features (denoted as `FG Re-sampler' and `BG Re-sampler', respectively).
To model their underlying dependencies, a relating component named \textit{Context Bridge} is involved to explicitly establish the connection between the two re-samplers.
An overview of the Dual Re-sampler architecture is provided in Figure~\ref{fig:dual_resampler}.

\vspace{3pt}

\noindent\textbf{FG Re-sampler.}
Inspired by the Perceiver architecture~\cite{alayrac2022flamingo}, the FG Re-sampler utilizes a set of learnable queries $\mathbf{q^{fg}}\in\mathbb{R}^{1\times N_q\times d}$ (where $N_q$ denotes the number of query tokens and $d$ the latent dimension) to extract the semantic information of foreground objects.
Specifically, given the text embedding of object labels $\mathbf{f^{obj}} \in \mathbb{R}^{N\times S \times d}$ (computed by a pre-trained CLIP encoder~\cite{radford2021learning} and $S$ denotes the sequence length), the FG Re-sampler comprises $L$ attention layers, each consisting of a Cross Attention (CA) block followed by a Feed Forward Network (FFN). 
Formally, the CA mechanism in FG Re-sampler is expressed as:
\begin{equation}\label{eq:cross_attention}
\text{CA}(\mathbf{q^{fg}}, \mathbf{f^{obj}}) = \text{Softmax}\left(\frac{\mathbf{Q}(\mathbf{\mathbf{q^{fg}}})\mathbf{K}(\mathbf{f^{obj}})^\top}{\sqrt{d}}\right)\mathbf{V}(\mathbf{f^{obj}}),
\end{equation}
where the output of each layer provides the query input for the subsequent layer, ultimately yielding the FG token $\mathbf{h^{fg}}\in \mathbb{R}^{N\times N_q\times d}$.

Beyond capturing semantic information, the grounding token $\mathbf{h^{bbox}} \in \mathbb{R}^{N\times 1\times d}$ encodes the spatial distribution of objects by mapping bounding box parameters through a Fourier Embedding module~\cite{mildenhall2021nerf}. 
The element-wise sum of $\mathbf{h^{fg}}$ and $\mathbf{h^{bbox}}$ then serves as the comprehensive FG representation passed on to the subsequent processing stages.

\vspace{3pt}

\noindent\textbf{Context Bridge (CB).}
To bridge FG and BG and improve coherence, we additionally introduce a learnable context query $\mathbf{q^{ctx}} \in \mathbb{R}^{1\times N_q \times d}$. 
Specifically, $\mathbf{q^{ctx}}$ is fed into a Self-Attention (SA) block together with the previously learned FG tokens (\textit{i.e.}, $\mathbf{h^{fg}}$ and $\mathbf{h^{bbox}}$) to capture cross-token relationships and contextual cues.
The resulting output is then added back to $\mathbf{q^{ctx}}$, yielding the final output of the Context Bridge, denoted as $\text{CB}(\mathbf{q^{ctx}})$.
Mathematically, $\text{CB}(\mathbf{q^{ctx}})$ can be expressed as
\begin{equation}
\label{eq:gsa}
\text{CB}(\mathbf{q^{ctx}}) = \mathbf{q^{ctx}}+\tanh (\gamma) \cdot \text{TS}(\text{SA}([\mathbf{q^{ctx}},\mathbf{h^{fg}}+\mathbf{h^{bbox}}])),
\end{equation}
where $\text{TS}(\cdot) $ is a Token Selection operation~\cite{li2023gligen} that retains only the context query from the SA output, $[\cdot]$ represents the concatenating operation and $\gamma$ is a learnable scalar.
By extracting the contextual cues of FG instances, $\mathbf{q^{ctx}}$ further enriches the context-aware processing of BG features.

\vspace{3pt}

\noindent\textbf{BG Re-sampler.}
Following the design of the FG Re-sampler, the BG Re-sampler also employs a learnable query $\mathbf{q^{bg}} \in \mathbb{R}^{1\times N_q\times d}$ to extract global semantics information from the textural description $\mathcal{P}$ for background synthesis.
However, unlike the FG Re-sampler, it additionally incorporates grounded context from FG objects through $\text{CB}({\mathbf{q^{ctx}}})$. 
Concretely, the concatenated query ($[\mathbf{q^{bg}}, \text{CB}({\mathbf{q^{ctx}}})]$) engages in the CA mechanism with the text embedding of $\mathcal{P}$ (denoted as $\mathbf{f^{all}}\in\mathbb{R}^{1\times S \times d}$).
As a result, the output BG token $\mathbf{h^{bg}}\in\mathbb{R}^{1\times2N_q\times d}$ encodes BG semantics with the awareness of FG context.

\vspace{3pt}

\noindent\textbf{Condition Embeddings.}
The final outputs of the Dual Re-sampler are the condition embeddings for the $N$ FG objects ($\{\mathbf{e^{fg}_{i}}\}_{i=1}^N$) and the BG ($\mathbf{e^{bg}}$).
These embeddings are constructed by concatenating the corresponding learned tokens, as follows:
\begin{align}
    \mathbf{e^{fg}_{i}} & = [\mathbf{h_{i}^{fg}}, \mathbf{h_i^{bbox}}, \texttt{cls}(\mathbf{c_{i}})] \in \mathbb{R}^{1\times(N_q+2)\times d} \\
    \mathbf{e^{bg}}     & = [\mathbf{h^{bg}}, \texttt{cls}(\mathcal{P})],
\end{align}
where $\mathbf{h_{i}^{fg}}$, $\mathbf{h_i^{bbox}}$ are the token slices pertaining to the $i$-th object, and $\texttt{cls}(\cdot)$ denotes the output $\texttt{cls}$ token produced by the CLIP text encoder.

%
%

\begin{figure*}[ht]
    \begin{center}
        \includegraphics[width=0.925\linewidth]{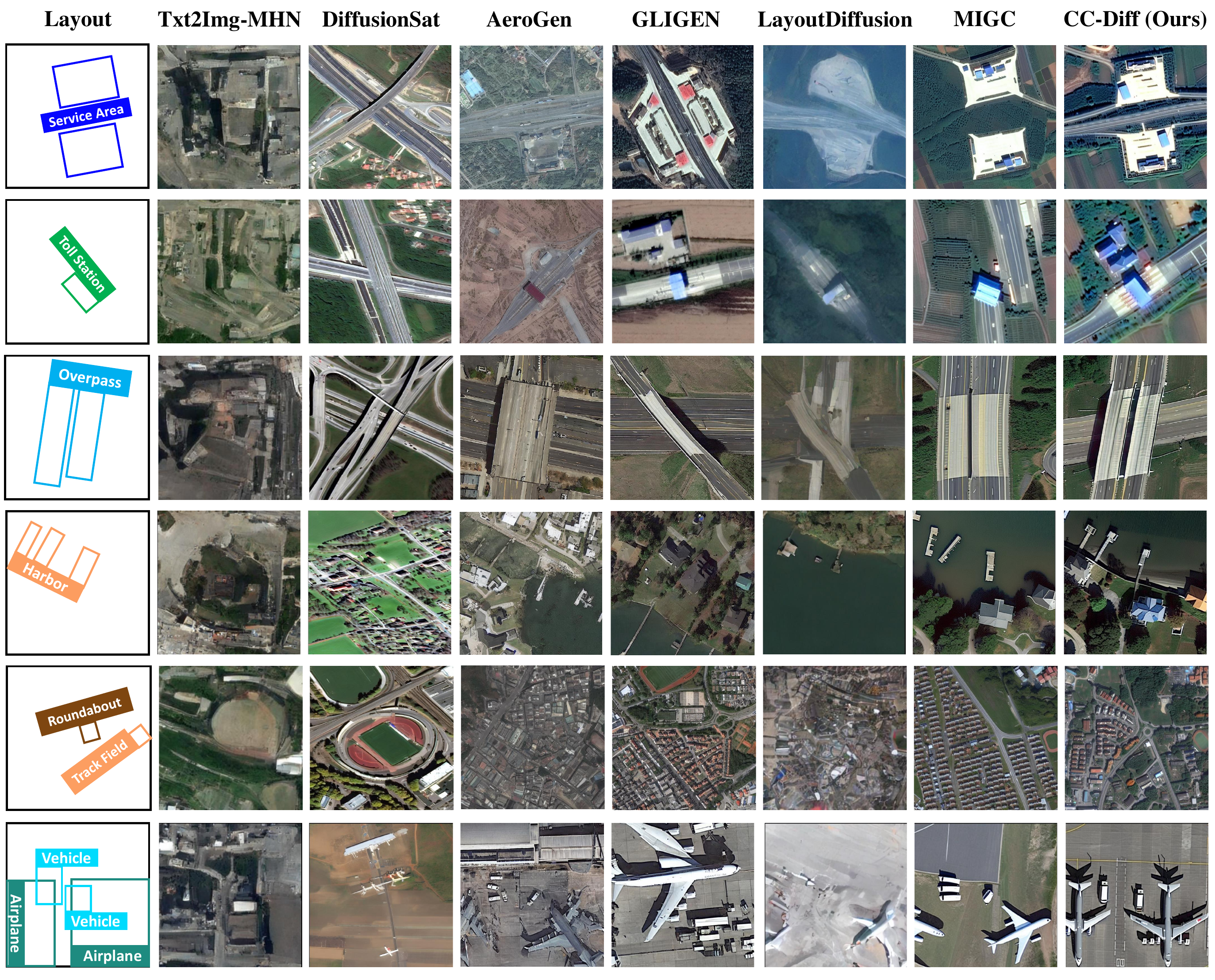}
    \end{center}
    \vspace{-10pt}
    \caption{Qualitative L2I results on DIOR-RSVG (top three rows) and DOTA (bottom three rows). CC-Diff not only synthesizes realistic foregrounds with accurate positioning but also generates more detailed backgrounds with stronger coherence to the foreground. Please zoom in for better details.}
    \label{fig:qual_res_RS}
\end{figure*}

\subsection{CGM for FG \& BG Generation}

Given the condition embeddings $\{\mathbf{e^{fg}_{i}}\}_{i=1}^N$ and ($\mathbf{e^{bg}}$) from the Dual Re-sampler, the Conditional Generation Module (CGM) synthesizes grounded FG instances in parallel and renders the BG using their integrated representation.
An illustration of CGM is provided in Figure~\ref{fig:cgm}.

\vspace{3pt}

\noindent \textbf{FG Instance Synthesis.}
For the $i$-th FG instance, the condition embedding $\mathbf{e^{fg}_i}$ acts as the controlling signal for the synthesis process, and the rendered feature map $\mathbf{R_i^{fg}}$ is obtained via the CA mechanism as follows
%
\begin{equation}\label{eq:ca_fg}
    \mathbf{R_i^{fg}} = \text{CA}(\mathbf{f}, \mathbf{e^{fg}_i})\odot{\mathbf{M_{i}}},
\end{equation}
where $\mathbf{f}$ denotes the incoming latent image feature to CGM from the previous layer in the LDM’s UNet, and $\mathbf{M_{i}}$ is the instance mask regulating the region of attention.
Notably, rather than using a binary mask, we adopt a non-uniform mask derived from a rotated \texttt{Sigmoid()} function~\cite{wang2024spotactor}, enabling smooth transitions between instances and their surrounding textures (see Appendix Sec.\ref{sec:sigmoid} for details).

\vspace{3pt}

\noindent \textbf{FG-aware BG Rendering.}
Unlike most existing methods~\cite{tang2024aerogen,zheng2023layoutdiffusion,zhou2024migc} that derive BG based on the image feature $\mathbf{f}$, we strengthen FG awareness by instead conducting BG synthesis based on the aggregation of all FG instance feature maps $\mathbf{R^{fused}} = \sum_{i=1}^{N}\mathbf{R^{fg}_{i}}$.
This design promotes a tighter coupling of FG and BG at the feature level, resulting in smoother transitions in the generated image (see Sec.~\ref{sec:ablation} for discussion).
Specifically, we also compute the BG feature $\mathbf{R^{bg}}$ via the \textit{FG-aware Attention} mechanism using CA, \textit{i.e.}, $\mathbf{R^{bg}} = \text{CA}(\mathbf{R^{fused}}, \mathbf{e^{bg}})$, and finally add $\mathbf{R^{fused}}$ to $\mathbf{R^{bg}}$ to obtain the final output of CGM.

\section{Experiments}
\label{sec:experiments}


\begin{table*}[!t]
\caption{Quantitative comparison of results on RS datasets DIOR-RSVG and DOTA. The detector for computing the YOLOScore struggles to detect most instances in images generated by GLIGEN, leading to notably low values (indicated with $\dagger$).}
\label{table:QuanRS}
\vspace{-5pt}
\centering
\resizebox{0.95\textwidth}{!}{
    \ra{1.15}
    \begin{tabular} {l rrrrr c l rrrrr}
    \Xhline{1.0pt}
    \multicolumn{6}{c}{DIOR-RSVG}                           &\phantom{a} & \multicolumn{6}{c}{DOTA} \\
    
    \cline{1-6}                                                    \cline{8-13}
    
    \multirow{2.3}{*}{Method} & \multicolumn{2}{c}{CLIPScore $\uparrow$} & \multirow{2.3}{*}{FID $\downarrow$} & \multicolumn{2}{c}{YOLOScore $\uparrow$} &\phantom{a} & 
    \multirow{2.3}{*}{Method} & \multicolumn{2}{c}{CLIPScore $\uparrow$} & \multirow{2.3}{*}{FID $\downarrow$} & \multicolumn{2}{c}{YOLOScore $\uparrow$} \\
    
    \cline{2-3} \cline{5-6} \cline{9-10} \cline{12-13} 
    
     & Local & Global & & \multicolumn{1}{c}{$\text{mAP}_{50}$} & \multicolumn{1}{c}{$\text{mAP}_{50-95}$} &\phantom{a} & & Local & Global & & \multicolumn{1}{c}{$\text{mAP}_{50}$} & \multicolumn{1}{c}{$\text{mAP}_{50-95}$} \\

    \cline{1-6}                                                    \cline{8-13}
   
    Txt2Img-MHN      & $18.91$ & $23.46$ & $123.84$ & $0.30$ & $0.08$  &\phantom{a} & Txt2Img-MHN      & $19.58$  & $25.99$  & $137.76$ & $0.02$ & $0.01$ \\
    DiffusionSat     & $19.84$ & \cellcolor{gray!20}$\bm{32.68}$ & $78.16$ & $0.80$ & $0.20$   &\phantom{a} & DiffusionSat     & $19.78$  & \cellcolor{gray!20}$\bm{31.61}$  & $65.19$ & $0.04$ & $0.01$ \\
    GLIGEN           & $20.55$ & $32.22$ & $73.02$ & $3.44^{\dagger}$ & $0.75^{\dagger}$   &\phantom{a} & GLIGEN           & $20.72$  & $29.98$  & $61.05$ & $0.25^{\dagger}$ & $0.07^{\dagger}$ \\
    AeroGen      & $20.28$ & $30.75$ & $74.90$ & $40.93$ & $21.29$  &\phantom{a} & AeroGen      & $21.55$  & $26.13$  & $55.02$ & $25.72$ & $12.47$ \\
    LayoutDiffusion  & $19.31$ & $30.65$ & $79.03$ & $56.92$ & $31.05$ &\phantom{a} & LayoutDiffusion  & $20.49$  & $27.67$  & $64.77$ & $28.28$ & $11.40$\\
    MIGC             & $21.59$ & $32.36$ & $79.93$ & $59.55$ & $31.16$ &\phantom{a} & MIGC             & $22.21$  & $30.96$  & $63.95$ & $35.43$ & $14.85$\\
    CC-Diff (Ours)     & \cellcolor{gray!20}$\bm{21.82}$ & $32.36$ & \cellcolor{gray!20}$\bm{70.68}$ & \cellcolor{gray!20}$\bm{68.40}$ & \cellcolor{gray!20}$\bm{41.92}$ &\phantom{a} & CC-Diff (Ours)  & \cellcolor{gray!20}$\bm{22.60}$  & $30.92$  & \cellcolor{gray!20}$\bm{47.72}$ & \cellcolor{gray!20}$\bm{45.09}$ & \cellcolor{gray!20}$\bm{21.83}$\\

    \Xhline{1.0pt}

    \end{tabular}
}
\end{table*}

%

\subsection{Experimental Settings}
\label{sec:experimental_settings}

\noindent \textbf{Datasets.}
Our experiments are conducted on two remote sensing (RS) datasets: \textbf{DIOR-RSVG}~\cite{zhan2023rsvg} with 17,402 images and the more challenging \textbf{DOTA}~\cite{xia2018dota} with 2,806 images. 
We also use \textbf{COCO2017}~\cite{lin2014coco} to assess the generalizability of CC-Diff in handling diverse object categories and complex attributes in natural scenes.
Please refer to Sec.~\ref{sec:dataset} of the Appendix for details on dataset preparation.

%



\begin{table}[tb]
\caption{Trainability ($\uparrow$) comparison on DIOR-RSVG and DOTA. 'Baseline' denotes accuracy with the unaugmented dataset. GLIGEN is excluded due to low detection rates of foreground instances in generated samples.}
\vspace{-7pt}
\centering
    \resizebox{1.0\columnwidth}{!}{
    \ra{1.15}
    \begin{tabular}{l r r r c r r r}
        \Xhline{1.0pt}
        \multirow{2.3}{*}{Method} & \multicolumn{3}{c}{DIOR-RSVG} & \phantom{a} & \multicolumn{3}{c}{DOTA} \\
        \cline{2-4} \cline{6-8}
        & mAP & $\text{mAP}_{50}$ & $\text{mAP}_{75}$ & \phantom{a} & mAP & $\text{mAP}_{50}$ & $\text{mAP}_{75}$  \\
        \hline
        Baseline        & 50.17 & 75.84 & 54.38 & & 35.53 & 62.10 & 35.83 \\
        Txt2Img-MHN     & 50.12 & 75.87 & 54.74 & & 35.91 & 62.53 & 36.43 \\
        DiffusionSat    & 49.95 & 75.59 & 55.26 & & 36.15 & 62.50 & 36.76 \\
        AeroGen         & 51.39 & 76.85 & 56.75 & & 36.65 & 63.15 & 36.96 \\
        LayoutDiffusion & 51.96 & 77.31 & 56.82 & & 35.15 & 61.54 & 35.18  \\
        MIGC            & 51.87 & 76.65 & 57.20 & & 35.93 & 62.36 & 36.32  \\
        CC-Diff (Ours)  & \cellcolor{gray!20}\textbf{52.18} & \cellcolor{gray!20}\textbf{77.39} & \cellcolor{gray!20}\textbf{57.59} & & \cellcolor{gray!20}\textbf{37.36} & \cellcolor{gray!20}\textbf{63.18} & \cellcolor{gray!20}\textbf{38.55} \\
        \Xhline{1.0pt}
     \end{tabular}
  }
\label{tab:RS_Trainability}
\end{table}

\vspace{3pt}
\noindent \textbf{Implementation Details.}
We adopt the pre-trained Stable Diffusion V1.4~\cite{rombach2022high} as the backbone for fine-tuning in all experiments.
The global text prompt for each image’s layout follows the rule-based protocol in~\cite{liu2024remoteclip}, while all learnable queries are randomly initialized using a standard normal distribution.
Following recent L2I efforts~\cite{zhou2024migc,zhou2024migc++}, we resize all images to $512\times 512$~\footnote{Images in DOTA are evenly segmented into the same resolution, without considering the integrity of whole objects.} and allow a maximum of six objects per image.
We train CC-Diff for $100$ epochs with a batch size of $320$ on $8 \times$ NVIDIA A800 GPUs, with a fixed learning rate of $1e^{-4}$.

\vspace{3pt}
\noindent \textbf{Benchmark Methods.}
We select both remote sensing (RS) and natural image generation methods as benchmarks.
For RS images, we adopt three recent controllable generation approaches: \textbf{Txt2Img-MHN}~\cite{xu2023txt2img}, \textbf{DiffusionSAT}~\cite{samar2024diffusionsat}, and \textbf{AeroGen}~\cite{tang2024aerogen}.
The first two rely on our previously described rule-based text prompts, while AeroGen additionally incorporates spatial layouts.
For a fair comparison of spatial controllability and to assess CC-Diff’s generalizability, we also include three state-of-the-art layout-to-image (L2I) methods for natural image generation: \textbf{GLIGEN}~\cite{li2023gligen}, \textbf{LayoutDiffusion}~\cite{zheng2023layoutdiffusion}, and \textbf{MIGC}~\cite{zhou2024migc}.

\vspace{3pt}
\noindent \textbf{Evaluation Metrics.}
Following~\cite{li2023gligen,zheng2023layoutdiffusion,zhou2024migc,chen2024geodiffusion}, we evaluate CC-Diff on three key criteria: 
\textbf{(1) Fidelity}, measured by the FID score~\cite{heusel2017fid} for perceptual quality; 
\textbf{(2) Faithfulness}, assessed using Global and Local CLIPScores~\cite{avrahami2023clipscore} for semantic consistency and YOLOScore~\cite{li2021yoloscore} for layout alignment; 
and \textbf{(3) Trainability}, evaluated by mean Average Precision (mAP) to test how well synthetic samples boost object detection accuracy.
See Appendix Sec.~\ref{sec:metrics} for further details.

\begin{figure*}[t]
    \begin{center}
        \includegraphics[width=0.8\linewidth]{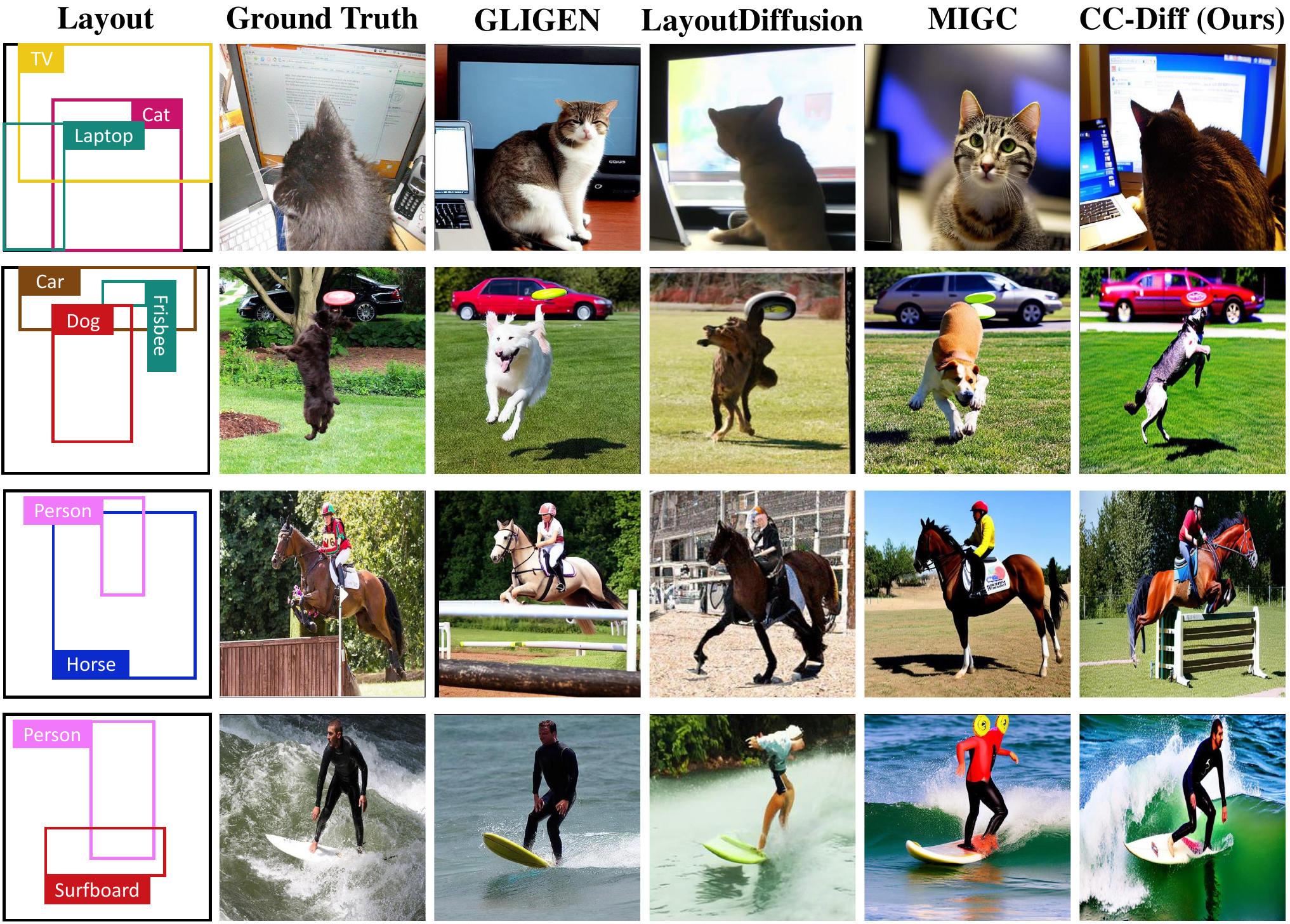}
    \end{center}
    \vspace{-10pt}
    \caption{Qualitative L2I results on the COCO dataset. In addition to generating realistic foregrounds, CC-Diff produces more plausible and coherent background integration. Please zoom in for better details.}
    \label{fig:qual_res_COCO}
\end{figure*}

\subsection{Results on RS Image Synthesis}
\label{sec:exp_RS}

We start by presenting experimental results on RS datasets, DIOR-RSVG and DOTA. 
To ensure a fair comparison, we use official checkpoints for all benchmark methods and fine-tune them on the respective training splits of each dataset.

\subsubsection{Qualitative Results on RS Datasets}
Figure~\ref{fig:qual_res_RS} presents a comparison of generated RS images.
CC-Diff not only synthesizes realistic foreground instances at varying scales, with accurate positioning and orientation consistent with the given layout, but also generates backgrounds with more intricate textures. 
More importantly, it establishes a significantly more coherent and plausible relationship between the foreground and background.
For example, in the first three cases, CC-Diff successfully renders the road going through the foreground instances, showing a reasonable association with the presence of the expressway service areas, toll station and overpass.

As for the benchmark methods, Txt2Img-MHN and DiffusionSat struggle to control the location of foreground objects, despite the inclusion of spatial information in global text prompts. 
Among the L2I approaches, MIGC achieves the best visual quality and semantic consistency. However, compared to CC-Diff, its rendered background lacks coherence with the foreground, resulting in an implausible landscape.
Please refer to Appendix Sec. \ref{sec:detailed_quali_results} for more results.

\subsubsection{Quantitative Results on RS Datasets}
\label{sec:QuanRS}

Quantitative results on RS datasets are shown in Table~\ref{table:QuanRS} and Table~\ref{tab:RS_Trainability}.
In this section, we present a comprehensive analysis from three perspectives including realism, faithfulness, and trainability, as outlined in Sec.~\ref{sec:experimental_settings}.

\noindent \textbf{Visual Fidelity.}
%
%
CC-Diff achieves leading FID scores of $70.68$ on DIOR-RSVG and $47.72$ on DOTA, outperforming the second-best methods by $2.34$ points (GLIGEN: $73.02$) and $7.30$ points (AeroGen: $55.02$), respectively. 
This clear performance edge can be attributed to CC-Diff’s effective synthesis of realistic instance textures and its enhanced alignment with the overall scene context.

\noindent \textbf{Semantic Faithfullness.}
CC-Diff demonstrates state-of-the-art performance in global and regional semantic alignment, as indicated by CLIPScore values. 
While DiffusionSat achieves a marginally higher Global CLIPScore ($32.68$ \textit{vs.} $32.36$), this is likely due to its larger RS-specific pre-training dataset~\cite{samar2024diffusionsat}. 
Additionally, CC-Diff shows a clear advantage in YOLOScore, confirming that instances are well recognized by the object detector, with accurate layout retention reflected in the high mAP values.

%
%
%

\noindent \textbf{Trainability.}
Following the data enhancement protocol in~\cite{chen2024geodiffusion}, we double the training samples using layout-based synthesis and assess detection accuracy with the expanded dataset.
As shown in Table~\ref{tab:RS_Trainability}, CC-Diff consistently delivers the highest accuracy gains, improving mAP by $2.01$ on DIOR-RSVG and $1.83$ on DOTA. Please refer to Appendix Sec. \ref{sec:train_ap} for additional results in fine-grained settings.
%

\subsection{Generalization to Natural Image Synthesis}

\subsubsection{Qualitative Results on COCO}
The comparison of generation results on the COCO dataset is shown in Figure~\ref{fig:qual_res_COCO}.
CC-Diff effectively synthesizes realistic foreground instances and coherent backgrounds, where the semantic connections to the foreground align well with the ground truth.
These results demonstrate the strong generalizability of CC-Diff beyond RS image generation.

\begin{table}[tb]
\caption{Quantitative comparison of results on COCO.}
\centering
\vspace{-7pt}
\resizebox{\columnwidth}{!}{
    \ra{1.05}
    \begin{tabular}{l rrrrr}
        \Xhline{1.0pt}
        \multirow{2.1}{*}{Method} & \multicolumn{2}{c}{CLIPScore $\uparrow$} & \multirow{2.1}{*}{FID $\downarrow$} & \multicolumn{2}{c}{YOLOScore $\uparrow$} \\
        \cline{2-3} \cline{5-6}  
    
        & Local & Global && \multicolumn{1}{c}{$\text{mAP}_{50}$} & \multicolumn{1}{c}{$\text{mAP}_{50-95}$} \\

        \hline
    
        GLIGEN          & $24.45$ & $30.60$ & \cellcolor{gray!20}$\bm{28.69}$ & $57.52$ & $35.84$ \\
        LayoutDiffusion & $23.15$ & $21.40$ & $37.26$ & $34.96$ & $17.95$ \\
        MIGC            & $24.75$ & $28.84$ & $34.31$ & \cellcolor{gray!20}$\bm{59.87}$ & $34.64$ \\
        CC-Diff (Ours)  & \cellcolor{gray!20}$\bm{24.88}$ & \cellcolor{gray!20}$\bm{31.45}$ & $30.35$ & $59.78$ & \cellcolor{gray!20}$\bm{36.71}$ \\
        
        \Xhline{1.0pt}
  \end{tabular}
  }
  \label{tab:QuanCOCO}
\end{table}

\begin{table}[tb]
\caption{Trainability ($\uparrow$) comparison on COCO.  `Baseline' denotes accuracy with the unaugmented dataset.}
\vspace{-7pt}
\centering
    \resizebox{0.7\columnwidth}{!}{
    \ra{1.15}
    \begin{tabular}{l r r r}
    
        \Xhline{1.0pt}
        
        \multirow{1.2}{*}{Method} 
        
        & mAP & $\text{mAP}_{50}$ & $\text{mAP}_{75}$   \\
        
        \hline
        
        Baseline        & 35.35 & 59.53 & 37.50 \\
        GLIGEN     & 37.51 & 61.18 & \cellcolor{gray!20}\textbf{39.95}  \\
        LayoutDiffusion & 36.39 & 59.81 & 38.65 \\
        MIGC            & 37.01 & 60.45 & 39.38   \\
        CC-Diff (Ours)  & \cellcolor{gray!20}\textbf{37.60} & \cellcolor{gray!20}\textbf{61.44} & 39.93  \\
        \Xhline{1.0pt}
     
     \end{tabular}
  }
\label{tab:COCO_Trainability}
\end{table}

\subsubsection{Quantitative Results on COCO}


\noindent \textbf{Visual Fidelity.}
As shown in Table~\ref{tab:QuanCOCO}, CC-Diff achieves an FID score of $30.35$ on COCO dataset. 
Although it is slightly lower than the seminal study GLIGEN~\cite{li2023gligen} by $1.66$ FID, CC-Diff outperforms MIGC ($34.31$) by $3.96$ and LayoutDiffusion ($37.26$) by $6.91$. 
This demonstrates the promising generalizability of CC-Diff to natural image datasets with more diverse and complex foreground attributes.

\noindent \textbf{Semantic Faithfullness.}
As indicated by the CLIPScore in Table~\ref{tab:QuanCOCO}, CC-Diff achieves the best semantic consistency performance at both local and global levels. 
While MIGC slightly outperforms CC-Diff by $0.09$ YOLOScore under $\text{mAP}{50}$ ($59.87$ \textit{vs.} $59.78$), CC-Diff achieves a more substantial improvement over all benchmarks under \text{mAP}$50$-$95$, demonstrating a stronger ability to preserve layout consistency across a broader range of threshold levels.

\noindent \textbf{Trainability.}
According to Table~\ref{tab:COCO_Trainability}, CC-Diff consistently improves object detection accuracy by incorporating synthetic augmented samples, achieving the largest accuracy gain of $2.25$ (from $35.35$ to $37.60$) under the most comprehensive metric (mAP). 
This demonstrates that the trainability of synthetic samples generated by CC-Diff can be effectively generalized from RS to natural images.

\begin{table}[tb]
\caption{Ablation on Context Bridge and FG-aware Attention.}
\label{tab:ablation_structure}
\centering
\vspace{-7pt}
    \resizebox{\columnwidth}{!}{
        \ra{1.15}
        \begin{tabular}{c c  r r r r r }
        
        \Xhline{1.0pt}
        
        \multirow{2.1}{*}{\makecell{Context \\ Bridge}} & \multirow{2.1}{*}{\makecell{FG-aware \\ Attention}} & \multicolumn{2}{c}{CLIPScore $\uparrow$} & \multirow{2.1}{*}{FID $\downarrow$} & \multicolumn{2}{c}
        {YOLOScore $\uparrow$} \\
    
        \cline{3-4} \cline{6-7}
    
         & & Local & Global && \multicolumn{1}{c}{$\text{mAP}_{50}$} & \multicolumn{1}{c}{$\text{mAP}_{50-95}$} \\
    
        \hline
        
        \ding{55}  & \ding{55} & $21.68$ & $32.33$ & $72.30$ & $63.08$ & $37.49$   \\
        \ding{55}  & \ding{51} & \cellcolor{gray!20}$\bm{21.90}$ & $32.32$ & $71.83$ & $64.98$ & $38.42$   \\
        \ding{51}  & \ding{55} & $21.76$ & $32.34$ & $71.28$ & $64.38$ & $39.22$   \\
        \ding{51}  & \ding{51} & $21.82$ & \cellcolor{gray!20}$\bm{32.36}$ & \cellcolor{gray!20}$\bm{70.68}$ & \cellcolor{gray!20}$\bm{68.40}$ & \cellcolor{gray!20}$\bm{41.92}$ \\
    
        \Xhline{1.0pt}

  \end{tabular}
  }
\end{table}

\begin{figure}[t]
  \centering
   \includegraphics[width=1.0\linewidth]{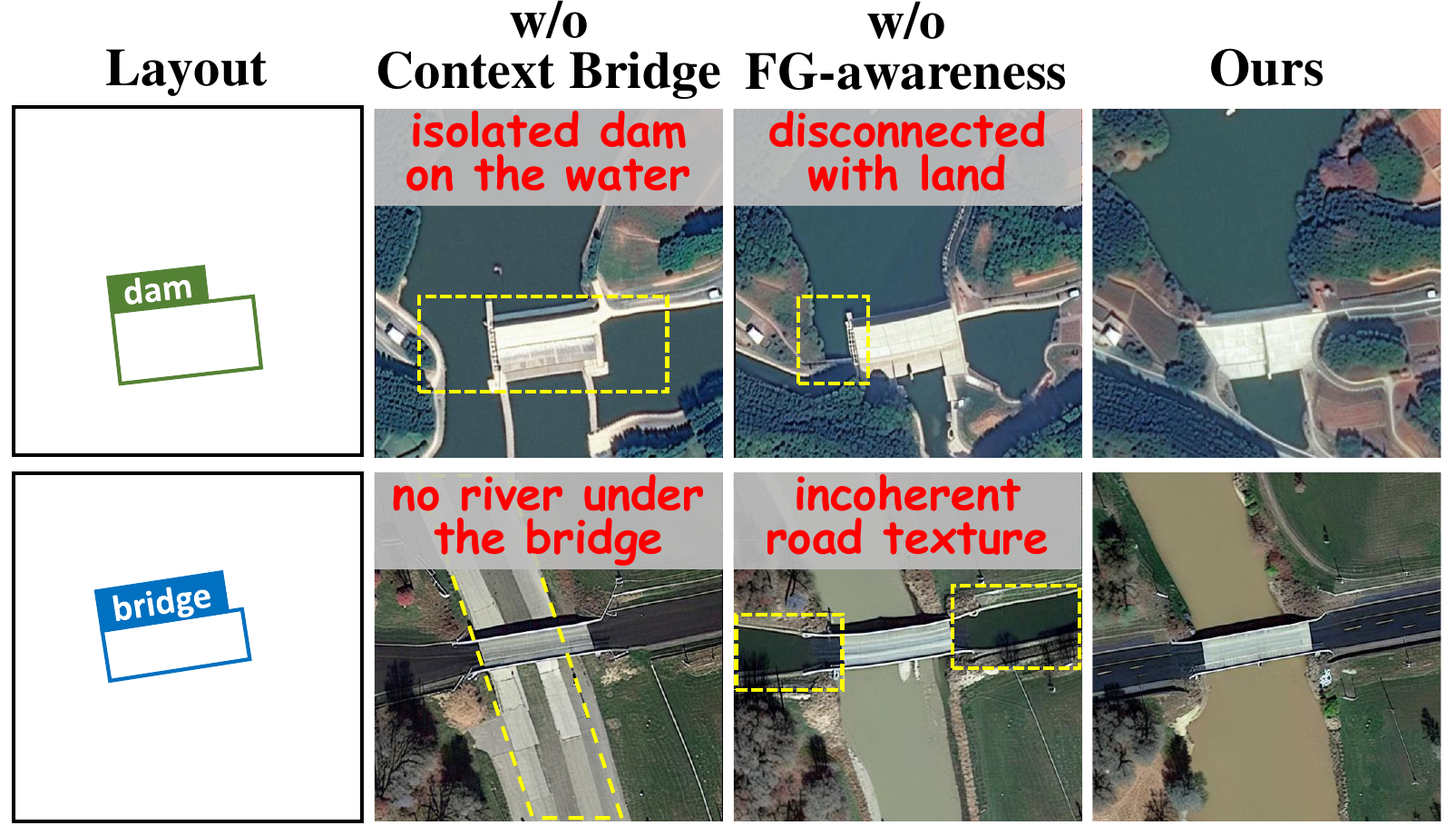}

   \caption{Qualitative results from various CC-Diff variants. Incoherent regions marked by dashed yellow boxes and labeled on top.}
   \label{fig:qual_abl}
\end{figure}

\subsection{Ablation Study}\label{sec:ablation}


\noindent \textbf{Conditional Generation Pipeline.}
%
%
%
We validate CC-Diff’s two core innovations, \textit{i.e.}, Dual Re-sampler and CGM, by separately removing the Context Bridge and FG-feature Attention mechanism to assess their individual contributions.
%
As shown in Table~\ref{tab:ablation_structure}, the Context Bridge significantly improves realism, with a 1.02 FID gain in minimal cases, and also benefits the Global CLIPScore (Row 1$^\text{st}$\&2$^\text{nd}$ \textit{vs.} Row 3$^\text{rd}$\&4$^\text{th}$). 
Meanwhile, introducing FG-awareness in background rendering raises both YOLOScore and FID (Row 1$^\text{st}$ \textit{vs.} 2$^\text{nd}$, Row 3$^\text{rd}$ \textit{vs.} 4$^\text{th}$). 
Furthermore, the qualitative results in Figure~\ref{fig:qual_abl} show that removing the Context Bridge (second column) disrupts the alignment between the background and its foreground context, whereas omitting FG-awareness (third column) introduces noticeably incoherent textures relative to foreground objects.
%
%
%
%


\noindent \textbf{LLM-assisted T2I.}
While synthetic RS images demonstrate strong trainability, achieving realistic layout guidance still requires extensive annotation.
To address this challenge, we leverage Large Language Models (LLMs) to generate plausible foreground layouts from the same rule-based textual descriptions~\cite{liu2024remoteclip} used to prompt Txt2Img-MHN and DiffusionSat in earlier experiments (Sec.~\ref{sec:exp_RS}).

\begin{figure}[t]
  \centering
   \includegraphics[width=1.0\linewidth]{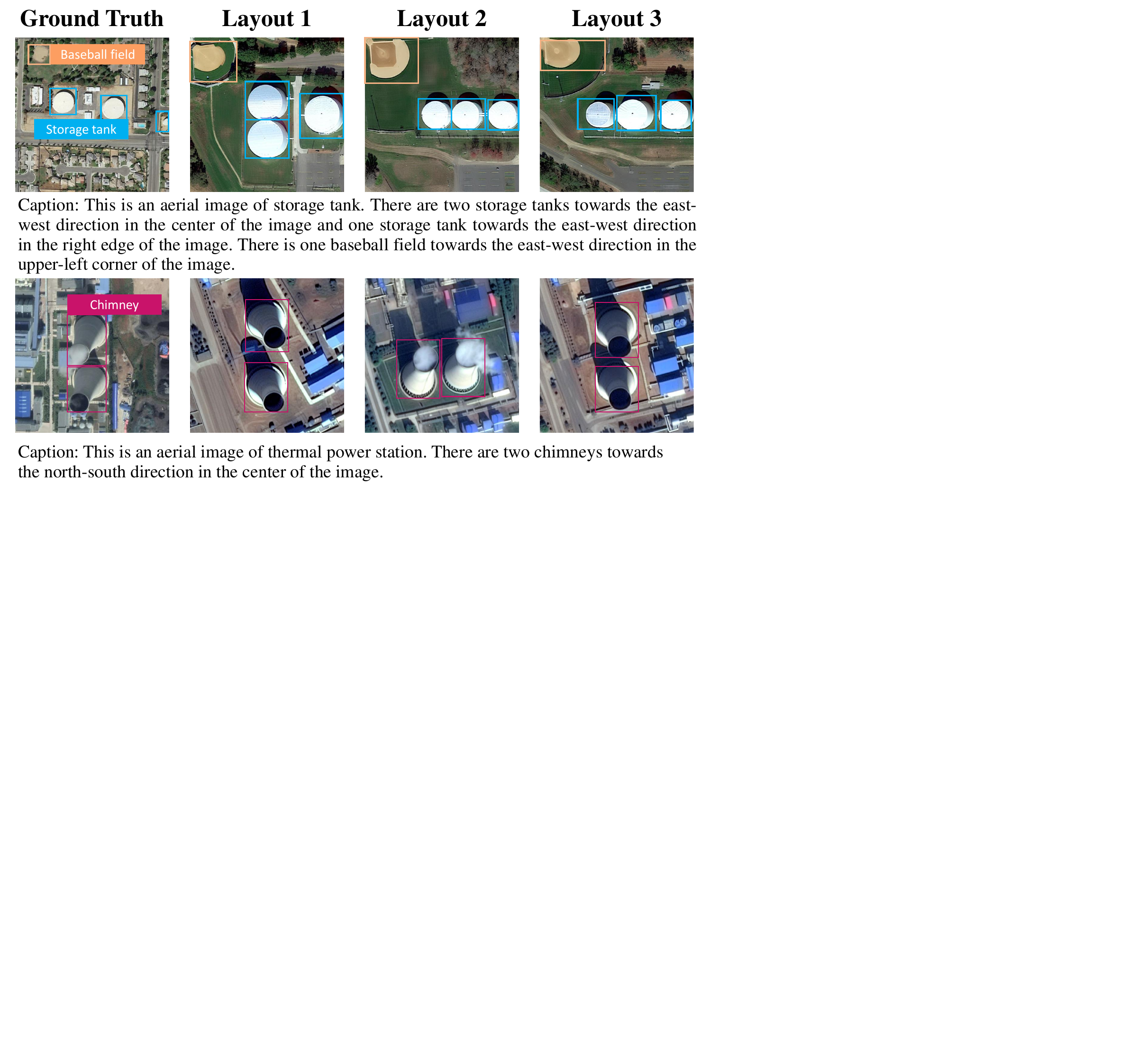}

   \caption{Illustration of various layouts generated by GPT-4 from the same rule-based caption (shown below), along with the corresponding images synthesized by CC-Diff.}
   \label{fig:qual_GPT}
\end{figure}

As shown in Table~\ref{tab:LLM_guided}, CC-Diff improves overall detection accuracy by $1.74$ mAP and consistently achieves the highest performance gains across all trainability settings. 
Additionally, CC-Diff demonstrates strong global semantic consistency, evidenced by a Global CLIPScore on par with Ground Truth and an FID score exceeding the second-best method by $6.10$ points ($76.71$ \textit{vs.} $70.61$).
Figure~\ref{fig:qual_GPT} further highlights CC-Diff’s adaptability to LLM-generated layout conditions, showcasing its promising zero-shot capability and the diversity enabled by this LLM-based T2I pipeline.


\begin{table}[tb]
\caption{($\uparrow$) performance on DIOR (layout generated with GPT-4o). GLIGEN is excluded due to low detection rates of foreground instances in generated samples.}
\vspace{-7pt}
\centering
\resizebox{\columnwidth}{!}{
    \ra{1.15}
    \begin{tabular}{l c c c c c }
        \Xhline{1.0pt}
        \multirow{2.1}{*}{Method} & \multicolumn{3}{c}{Trainability} & \multirow{2.1}{*}{\makecell{CLIPScore \\ Global}} & \multirow{2.1}{*}{FID} \\
        \cline{2-4}
         & mAP & $\text{mAP}_{50}$ & $\text{mAP}_{75}$ &  & \\
        \hline
        Base            & 50.17 & 75.84 & 54.38 & 31.70 & -  \\
        Txt2Img-MHN & 50.24 & 75.89 & 54.86 & 20.18 & 184.91   \\
        DiffusionSat & 49.99 & 75.49 & 54.31 & 32.64 & 79.33   \\
        AeroGen & 50.94 & 76.03 & 55.73 & 30.94 & 76.71   \\
        LayoutDiffusion & 50.96 & 76.46 & 56.17 & 30.16 & 77.48   \\
        MIGC            & 51.35 & 76.49 & 56.40 & \cellcolor{gray!20}\textbf{33.13} & 83.58   \\
        CC-Diff (Ours)  & \cellcolor{gray!20}\textbf{51.91} & \cellcolor{gray!20}\textbf{76.80} & \cellcolor{gray!20}\textbf{57.84} & 32.67 & \cellcolor{gray!20}\textbf{70.61}  \\
        \Xhline{1.0pt}
    \end{tabular}
  }
  \label{tab:LLM_guided}
\end{table}

\section{Conclusion}

Existing image generation methods often overlook the coherence between foreground and background, yet this is crucial for generating plausible RS images. 
To address this, we propose CC-Diff, an L2I method that focuses on rendering intricate background textures while ensuring a contextually coherent connection to foreground instances. 
By employing a sequential generation pipeline, CC-Diff conceptually models the interdependence of foreground and background, utilizing specific queries to capture fine-grained background features and their relationships. 
Experimental results confirm that CC-Diff can generate visually plausible images across both RS and natural domains, which also exhibit promising trainability on object detection.

\label{sec:conclusion}
{
    \small
    \bibliographystyle{ieeenat_fullname}
    \bibliography{main}
}

\clearpage
\setcounter{page}{1}
\maketitlesupplementary

Due to space constraints in the main submission, we provide the detailed implementation and explanation of the proposed CC-Diff in this Appendix.
Specifically, we present the implementation details of CC-Diff, including the definition of the bounding box orientation, the rule-based protocol for constructing the global text prompt $\mathcal{P}$, the implementation of instance mask and the setting details of datasets and metrics used in the experiments.
Additionally, we include further experimental results to underscore the effectiveness of CC-Diff.

\section{Definition of the Bounding Box Orientation}
\label{sec:def_angle}

\begin{figure}[t]
\centering
    \includegraphics[width=1.0\linewidth]{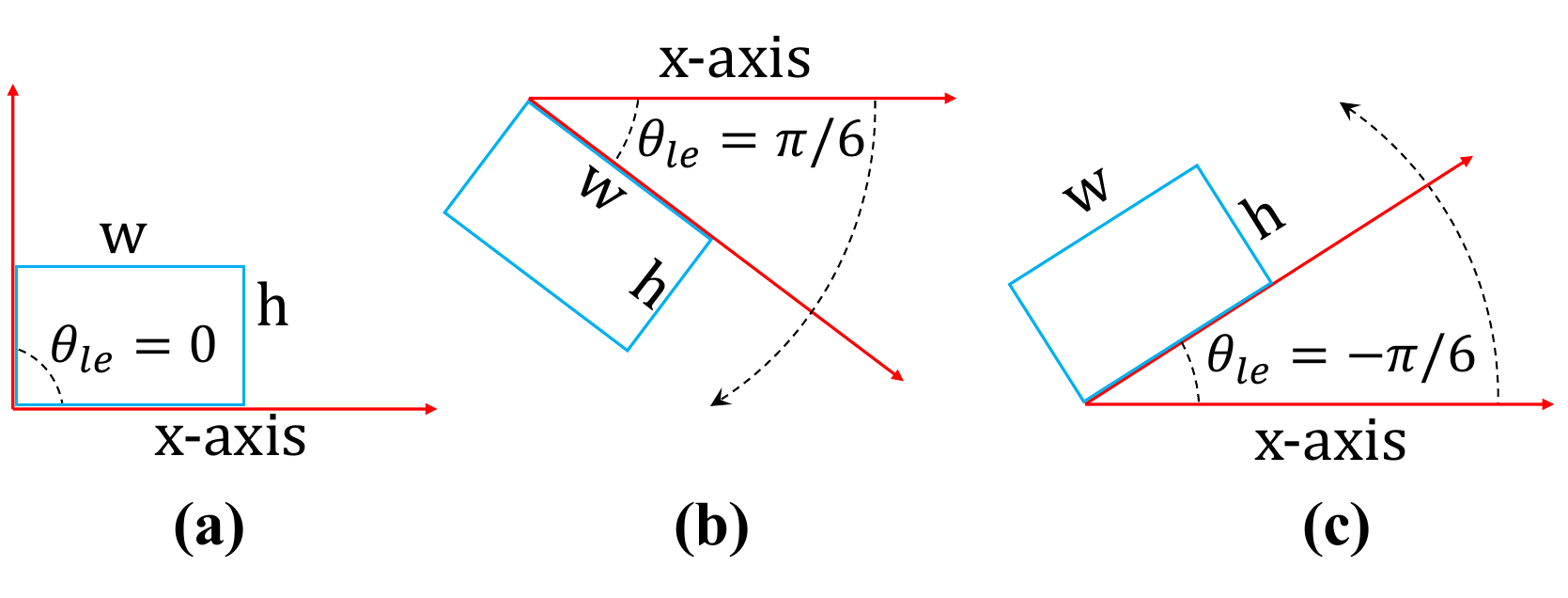}
    \caption{Definition of oriented bounding box (OBB) angles.}
    \label{fig:le90}
\end{figure}

Unlike natural images, where horizontal bounding boxes (HBB) are commonly used to delineate object contours, RS images require additional angular information to capture object orientation. 
This necessitates the use of oriented bounding boxes (OBB), which extend HBB by incorporating a rotation angle (as shown in Figure~\ref{fig:le90} (a)).

Given the existence of multiple conventions for defining the angular component of OBBs, we adopt the `long edge $90^\circ$ (le90)' definition throughout this study. 
Under this convention, an OBB is represented as $(x, y, w, h, \theta)$, where $(x, y)$ indicates the bounding box's center, $w$ and $h$ correspond to width and height of the box, and $\theta$ specifies the angle of rotation.

Specifically, the angle $\theta$ is measured between the longer edge of the bounding box and the positive x-axis, with clockwise rotations taken as positive and counterclockwise as negative (see Figure~\ref{fig:le90} (b) and (c)). 
This angle is confined to the range $[-90^\circ, 90^\circ)$ in degrees, which, in our experiments, is expressed in radians as $[-\pi/2, \pi/2)$.


\section{Construction of the Global Text Description and the GPT Prompt}
\label{sec:text_prompt}

\noindent \textbf{Global Text Description.}  The captions employed in recent Text-to-Image synthesis methods for remote sensing (RS) (e.g., Txt2Img-MHN~\cite{xu2023txt2img} and DiffusionSat~\cite{samar2024diffusionsat}) primarily describe the quantity and categories of foreground objects, often neglecting their spatial arrangement within the RS image.
To address this limitation and incorporate spatial guidance into the text prompt, we adopt a rule-based protocol from~\cite{liu2024remoteclip} to generate artificial descriptions, denoted as $\mathcal{P}$ in the main submission, that capture the spatial semantics of the scene.

\begin{figure}[th]
  \centering
   \includegraphics[width=1.0\linewidth]{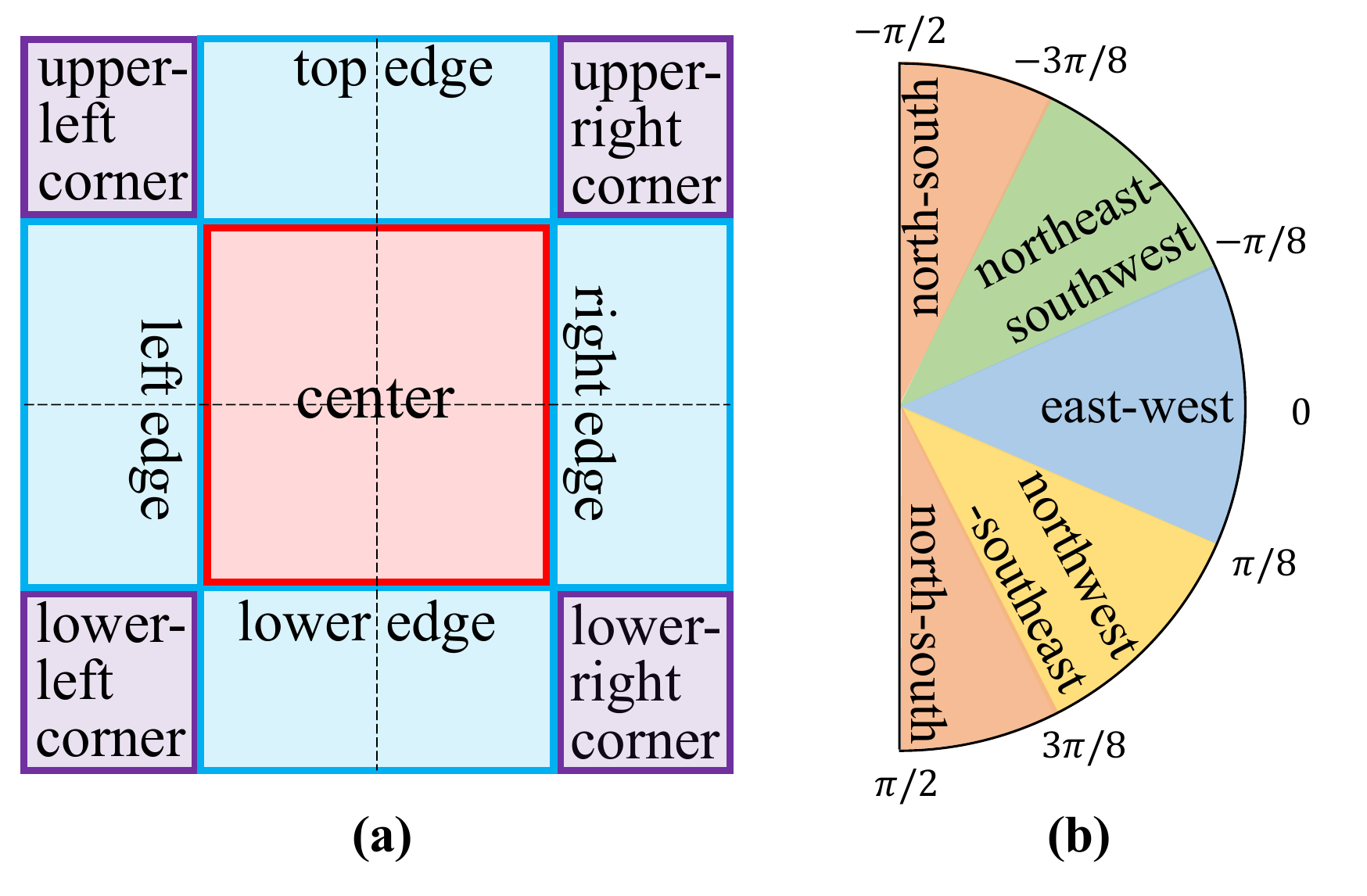}
   \vspace{-20pt}
   \caption{Illustration of (a) image space division and (b) object orientation definition for global textual description construction.}
   \label{fig:caption_illustration}
\end{figure}

As shown in Figure~\ref{fig:caption_illustration}, to efficiently incorporate spatial information into the text, this method divides the RS image into $K \times K$ blocks ($K=4$ for the example in Figure~\ref{fig:caption_illustration}), defining the $\lfloor K/2 \rfloor \times \lfloor K/2 \rfloor$ blocks at the center of the image as the `central region'. 
The remaining blocks are named based on their relative positions to the central region (\textit{e.g.}, `upper', `lower-right', \textit{etc.}). 
Following~\cite{lu2017rsicd}, the text description for a single image is constructed as follows:

\begin{enumerate}
    \item The first sentence offers a general description of the RS image to be generated. It follows the template: \texttt{`This is a remote sensing/an aerial image of <img\_cls>.'}, where \texttt{<img\_cls>} represents the class label of the RS image, obtained using the open-source classification model presented in~\cite{chen2024rsmamba}.

    \item Objects with their centers located in the central region are described first, with the class prioritized by descending instance count. The description for each type of foreground object follows the template: \texttt{`There is/are <obj\_num> of <obj\_cls> towards the <obj\_ori> direction in <obj\_pos>.'} Here, \texttt{<obj\_num>} represents the number of instances, \texttt{<obj\_cls>} specifies the object class, \texttt{<obj\_ori>} indicates the orientation, and \texttt{<obj\_pos>} denotes the block-level location within the image.

    \item For objects located outside the central region, the same template and prioritization are applied. 
    
    \item The total number of sentences in the prompt is capped at 5, with any remaining objects excluded from the description.
\end{enumerate}
Please refer to the sample captions shown in Figure~\ref{fig:qual_GPT} of the main submission, which serve as the global text descriptions used across all experiments.

\noindent \textbf{GPT Prompt.} 
We use GPT-4o~\cite{openai2024chatgpt} as the LLM for layout planning based on the given text prompt.
Inspired by the practice in~\cite{qu2023layoutllm,cho2023visual,kwon2024zero,wang2024divide,phung2024grounded,lian2024grounded}, the prompt consists of two main sections: \texttt{Instruction} and \texttt{Context Examples}.
In the \texttt{Instruction} section, we specify the task setting, constraints, and GPT's role. This includes details about the OBB format and the definition and significance of each of its components.
For the \texttt{Context Examples}, captions and layout pairs are retrieved using a reference image retrieval approach, selecting five samples with high semantic similarity to the query text description. These captions and layouts are incorporated into the prompt to define the expected input-output relationship for GPT, without including the original images themselves.
We provide an example of the GPT prompt as follows.

The instruction section of prompt mainly contains following parts:
\begin{enumerate}
    \item Task setting: 
    \begin{tcolorbox}
        You are an intelligent research assistant. I will provide the caption of an aerial image captured by a satellite. Your task is to:

        \begin{enumerate}
        \item Identify the object categories mentioned in the caption.
        \item Count the number of instances for each category.
        \item Generate an oriented bounding box (OBB) for each instance in the format: (object name, [center x, center y, width, height, rotation angle]).
        \end{enumerate}  
    \end{tcolorbox}
    \item Constraints:
    \begin{tcolorbox}
        \begin{itemize}
            \item[-] Image size is 512x512, with the top-left at [0, 0] and the bottom-right at [512, 512].
            \item[-] The width must always be greater than the height.
            \item[-] The rotation angle is the angle between the longer edge (width) and the positive x-axis, measured in degrees within [-90, 90]. 
            \item[-] Bounding boxes must stay entirely within the image boundaries.
            \item[-] Do not include objects not mentioned in the caption. 
        \end{itemize}
    \end{tcolorbox}
    \item GPT's role:
    \begin{tcolorbox}
        Validate that all bounding boxes meet the width $>$ height and boundary conditions. If necessary, make reasonable assumptions for object layout based on common aerial imagery.

        Please refer to the example below for the desired output format.
    \end{tcolorbox}
    
\end{enumerate}
We select the five examples with the highest similarity to the given caption as the in-context examples. An example is shown as follows:
\begin{tcolorbox}
    Example $\sharp$1:\\
    $<$Input Caption$>$\\
    This is an aerial image of airplane. There are three airplanes, two towards the northwest-southeast direction, one towards the northeast-southwest direction in the center of the image.\\
    $<$Output Bounding Boxes$>$\\
    airplane: [247, 221, 121, 112, 30]\\
    airplane: [306, 357, 110, 105, 33]\\
    airplane: [207, 336, 120, 112, -36]\\
\end{tcolorbox}

\section{Implementation of the Instance Mask}
\label{sec:sigmoid}
In the process of FG instance synthesis, the output of each cross-attention block is regulated via multiplying by an instance mask $\mathbf{M}$ by element which is converted from bbox $\mathbf{b}=[x_c, y_c, w, h, \theta]$. Unlike most existing methods adopting binary masks, we resort to a non-uniform implementation using the \texttt{Sigmoid()} function, which can be written as
\begin{equation}
\begin{split}
    \text{Sigmoid}(x, y) &= \frac{1}{1 + \text{exp}(-1+\frac{{(x^{\prime}-{\mu}_{1})}^{2}}{{\sigma}_{1}^2}+\frac{{(y^{\prime}-{\mu}_{2})}^{2}}{{\sigma}_{2}^2})} \\
    \begin{pmatrix}
        x^{\prime} \\ y^{\prime}
    \end{pmatrix}
    &=
    \begin{pmatrix}
    \cos{\theta} & \sin{\theta} \\
    -\sin{\theta} & \cos{\theta}
    \end{pmatrix}
    \begin{pmatrix}
        x \\ y
    \end{pmatrix}
\end{split}
\end{equation}
where $\mu_1$, $\mu_2$ represent the center of the oriented bounding box ($x_c, y_c$),  $\sigma_1 = w/2$, $\sigma_2 = h/2$ and $x^{\prime}, y^{\prime}$ are the results of rotating the grid coordinates ($x, y$) counterclockwise by $\theta$ angle.

\section{Dataset Details}
\label{sec:dataset}
Following datasets are used in experiments:
%
\begin{itemize}
    \item \textbf{DIOR-RSVG}~\cite{zhan2023rsvg} comprises $17,402$ RS images with a broad spectrum of landscape scales. The number of objects per category is by default restricted to a maximum of $5$, positioning DIOR-RSVG as a controlled baseline for evaluation.

    \item \textbf{DOTA}~\cite{xia2018dota} is a widely used benchmark dataset for RS object detection, containing $2,806$ images of varying sizes ranging from $800$ to $4,000$ pixels. It includes $15$ object categories, with no upper limit on the number of objects per image, making DOTA a challenging and practical dataset for our experiments.

    \item \textbf{COCO2017}~\cite{lin2014coco} is a standard benchmark for natural image generation. We employ it to evaluate the generalizability of CC-Diff in handling diverse object categories and complex attributes in natural scenes.
\end{itemize}

\section{Evaluation Metrics Setting}
\label{sec:metrics}
Following~\cite{li2023gligen,zheng2023layoutdiffusion,zhou2024migc,chen2024geodiffusion}, we evaluate the performance of CC-Diff across three main aspects:
\begin{itemize}
    \item \textbf{Fidelity}: Synthesis results should appear visually plausible. We use the FID score~\cite{heusel2017fid} to evaluate perceptual quality, capturing texture realism and contextual coherence.

    \item \textbf{Faithfulness}: Generated images are expected to align with the provided prompt, with global and local semantic consistency assessed by the Global and Local CLIPScores~\cite{avrahami2023clipscore}.
    The YOLOScore~\cite{li2021yoloscore} computed by a YOLOv8 detector~\cite{yolov8_ultralytics} further evaluates the alignment of layout for input and output.

    \item \textbf{Trainability}: We examine the potential of considering synthetic images as augmented samples for improving the accuracy of object detection. The standard mean Average Precision (mAP) metric is used for evaluation.
\end{itemize}

\section{Additional Qualitative Results}
\label{sec:detailed_quali_results}

\begin{figure*}[!t]
    \begin{center}
        \includegraphics[width=1\linewidth]{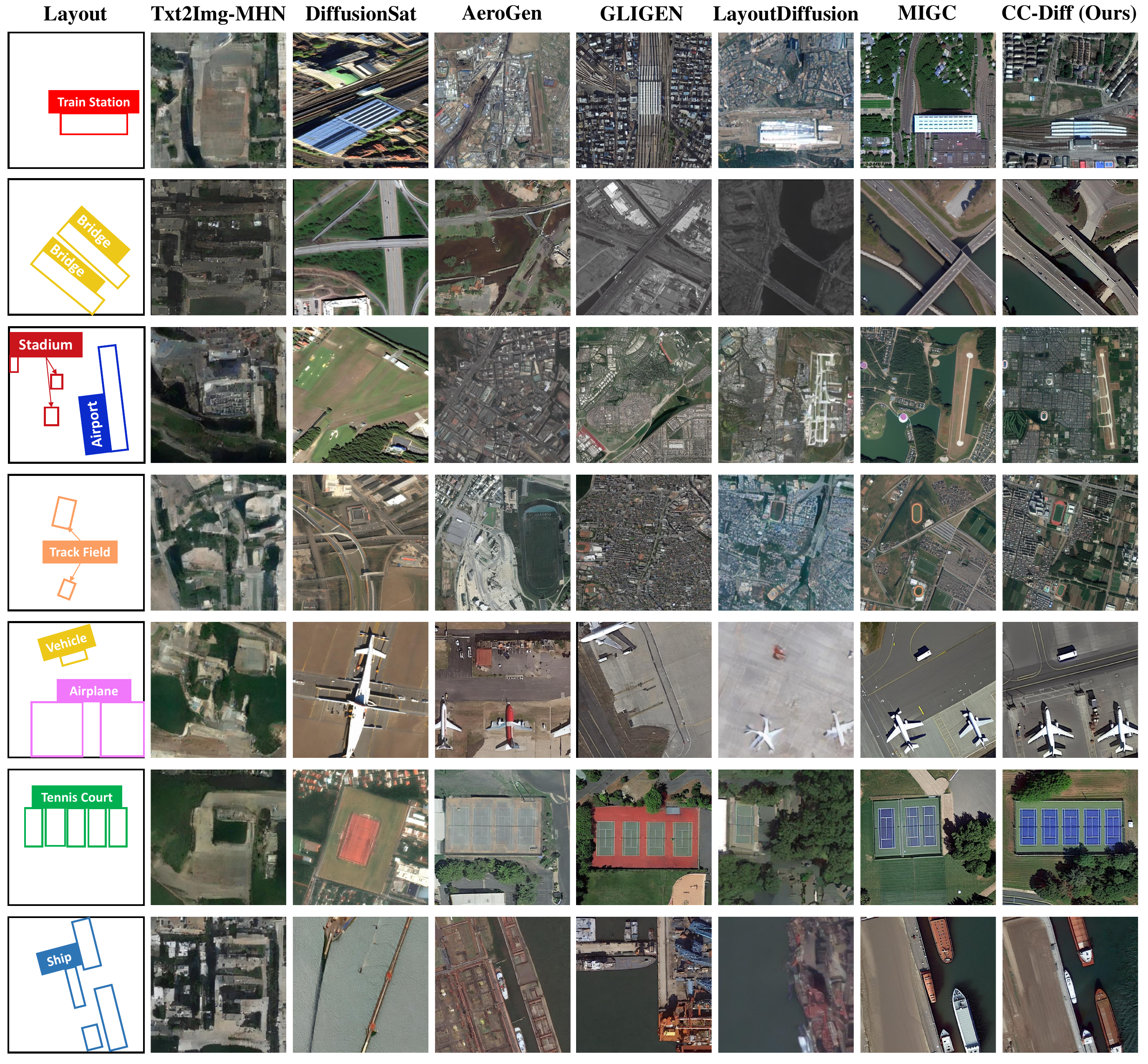}
    \end{center}
    \vspace{-10pt}
    \caption{Additional qualitative results are presented for DIOR-RSVG and DOTA. The first three rows highlight CC-Diff's ability to generate detailed backgrounds that exhibit strong coherence with the foreground. The middle three rows showcase its capability to synthesize images with complex backgrounds, while the last three rows demonstrate its effectiveness in generating scenes with multiple instances.}
    \label{fig:detail_qual_rs}
\end{figure*}

\begin{figure*}[!t]
    \begin{center}
        \includegraphics[width=0.910\linewidth]{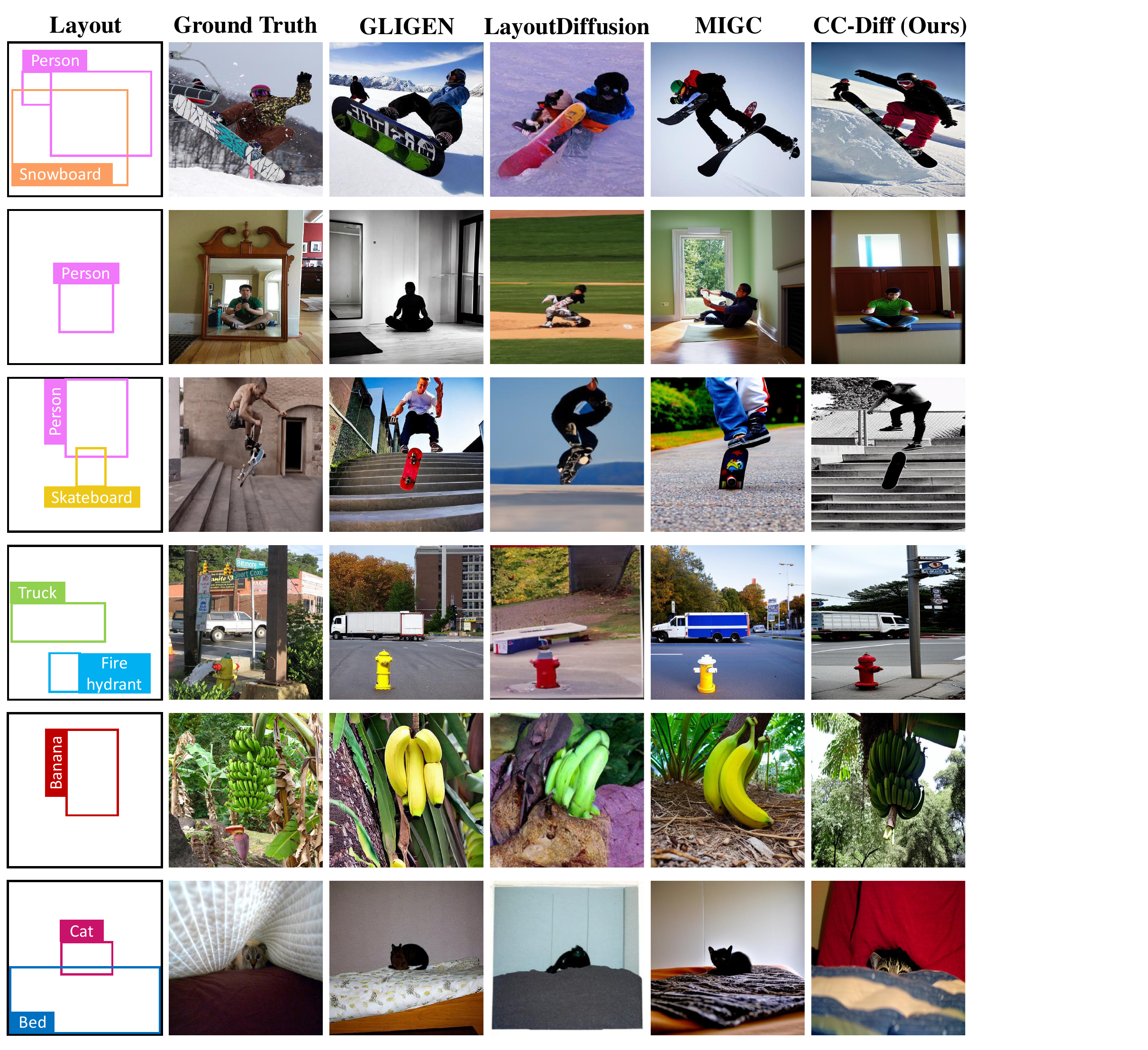}
    \end{center}
    \vspace{-15pt}
    \caption{Additional qualitative results on COCO. Beyond generating realistic foreground instances, CC-Diff demonstrates enhanced coherence and establishes more plausible relationships between the foreground and background.}
    \label{fig:detail_qual_coco}
\end{figure*}

Extra generation results on RS dataset and COCO are presented in Figure~\ref{fig:detail_qual_rs} and Figure~\ref{fig:detail_qual_coco}, respectively.

\section{Trainability AP}
\label{sec:train_ap}

The detailed trainability results are provided in Tables~\ref{table:train_ap_dior},~\ref{table:train_ap_dior_gpt}, and~\ref{table:train_ap_dota}, where the accuracy improvements for each individual object class are presented to facilitate a more thorough analysis.

From Table \ref{table:train_ap_dior}, it is evident that CC-Diff outperforms across multiple categories, particularly in complex and diverse scenarios. It achieves the highest scores in categories such as golf field ($54.65$), ground track field ($69.75$), train station ($28.63$), basketball court ($56.78$). These results underscore CC-Diff's strong generalization capabilities and its effectiveness in handling a variety of scenarios within the DIOR-RSVG dataset, demonstrating its potential for remote sensing image analysis. This ability is further reflected in the DIOR dataset with layout prompts generated by GPT-4o (Table \ref{table:train_ap_dior_gpt}) and the DOTA dataset for RS object detection (Table \ref{table:train_ap_dota}).

\begin{table*}[!t]
\caption{Detailed trainability results (measured by average precision) on DIOR-RSVG.}
\label{table:train_ap_dior}
\vspace{-5pt}
\centering
\resizebox{0.95\textwidth}{!}{
    \ra{1.15}
    \begin{tabular} {l ccccc ccccc}
    \Xhline{1.0pt}
    
    Method &  vehicle & chimney & golf field & Expressway-toll-station & stadium & ground track field & windmill & train station & harbor & overpass \\
    
    \hline
    Baseline         & 43.50  & 70.07  & 48.16  & 56.43  & 70.95  & 68.32  & 51.34  & 22.57  & 7.76  & 36.91 \\
    Txt2Img-MHN      & \textbf{43.61}  & 67.67  & 45.73  & 55.06  & 71.44  & 68.70  & 51.30  & 23.87  & 9.52  & 38.25 \\
    DiffusionSat     & 43.43  & 67.76  & 47.94  & 55.61  & 69.94  & 68.06  & 51.47  & 23.12  & 6.85  & 36.84 \\
    AeroGen          & 42.46  & 69.91  & 50.37  & 56.01  & 72.41  & 68.86  & 50.94  & 26.20  & 8.17  & \textbf{40.38} \\
    LayoutDiffusion  & 42.68  & 68.50  & 50.39  & \textbf{57.69}  & \textbf{73.30}  & 69.57  & \textbf{52.38}  & 28.23  & 9.20  & 40.33 \\
    MIGC             & 43.38  & 70.39  & 53.58  & 55.50  & 72.15  & 69.52  & 50.79  & 25.93  & \textbf{10.34} & 39.52 \\
    CC-Diff (Ours)   & 42.84  & \textbf{70.77}  & \textbf{54.65}  & 54.72  & 73.15  & \textbf{69.75}  & 51.83  & \textbf{28.63}  & 10.21 & 39.69 \\

    \Xhline{1.0pt}

    \end{tabular}
}
\end{table*}

\begin{table*}[!t]
\vspace{-5pt}
\centering
\resizebox{0.95\textwidth}{!}{
    \ra{1.15}
    \begin{tabular} {l ccccc ccccc}
    \Xhline{1.0pt}
    
    Method & baseball field & tennis court & bridge & basketball court & airplane & ship & storage tank & Expressway-Service-area & airport & dam \\
    
    \hline
    Baseline         & 76.81  & 54.63  & 25.97  & 54.05  & 68.18  & 51.77  & 80.22  & 42.15  & 44.17  & 29.38  \\
    Txt2Img-MHN      & 76.73  & 51.81  & 27.51  & 54.45  & 69.41  & 50.90  & 79.95  & 43.80  & 41.06  & 31.56  \\
    DiffusionSat     & 76.34  & 54.31  & 27.23  & 53.59  & 69.21  & 51.35  & 79.27  & 43.54  & 41.86  & 31.30 \\
    AeroGen          & \textbf{77.62}  & 54.86  & 28.07  & 54.61  & 66.96  & \textbf{54.04}  & \textbf{81.33}  & 47.84  & 44.78  & 31.89 \\
    LayoutDiffusion  & 76.83  & 54.34  & \textbf{30.36}  & 54.67  & 68.97  & 53.54  & 79.34  & \textbf{49.94}  & 46.85  & 32.05 \\
    MIGC             & 76.76  & 54.09  & 28.94  & 56.68  & \textbf{70.05}  & 52.91  & 80.57  & 46.22  & \textbf{47.57}  & \textbf{32.51} \\
    CC-Diff (Ours)   & 76.00  & \textbf{56.09}  & 28.92  & \textbf{56.78}  & 68.91  & 53.76  & 80.11  & 48.48  & 46.79  & 31.66 \\

    \Xhline{1.0pt}

    \end{tabular}
}
\end{table*}

\begin{table*}[!t]
\caption{Detailed trainability results (measured by average precision) on the DIOR dataset (layout generated using GPT-4o).}
\label{table:train_ap_dior_gpt}
\vspace{-5pt}
\centering
\resizebox{0.95\textwidth}{!}{
    \ra{1.15}
    \begin{tabular} {l ccccc ccccc}
    \Xhline{1.0pt}
    
    Method & vehicle & chimney & golf field & Expressway-toll-station & stadium & ground track field & windmill & train station & harbor & overpass \\
    
    \hline
    Baseline         & 43.50  & 70.07  & 48.16  & 56.43  & 70.95  & 68.32  & 51.34  & 22.57  & 7.76  & 36.91  \\
    Txt2Img-MHN      & 43.54  & 68.77  & 46.71  & \textbf{56.67}  & 71.49  & 69.06  & 50.73  & 24.00  & 8.91  & 37.21  \\
    DiffusionSat     & \textbf{43.65}  & 68.95  & 44.92  & 55.60  & 70.86  & 68.47  & 51.34  & 23.82  & 8.94  & 37.92 \\
    AeroGen          & 42.90  & 69.26  & 50.91  & 55.40  & 72.43  & 68.99  & 51.20  & 24.63  & 8.43  & \textbf{40.22}  \\
    LayoutDiffusion  & 42.91  & 67.57  & 49.34  & 55.06  & 73.24  & 68.74  & \textbf{52.43}  & \textbf{25.86}  & 6.22  & 38.13 \\
    MIGC             & 41.84  & 70.28  & 52.76  & 55.93  & 73.34  & \textbf{69.95}  & 51.54  & 24.00  & 9.16  & 37.14 \\
    CC-Diff (Ours)   & 42.82  & \textbf{71.46}  & \textbf{56.93} & 54.74  & \textbf{73.93}  & 68.92  & 50.95  & 24.75  & \textbf{10.03} & 38.06 \\

    \Xhline{1.0pt}

    \end{tabular}
}
\end{table*}

\begin{table*}[!t]
\vspace{-5pt}
\centering
\resizebox{0.95\textwidth}{!}{
    \ra{1.15}
    \begin{tabular} {l ccccc ccccc}
    \Xhline{1.0pt}
    
    Method & baseball field & tennis court & bridge & basketball court & airplane & ship & storage tank & Expressway-Service-area & airport & dam \\
    
    \hline
    Baseline         & 76.81  & 54.63  & 25.97  & 54.05  & 68.18  & 51.77  & 80.22  & 42.15  & 44.17  & 29.38  \\
    Txt2Img-MHN      & 76.72  & 52.88  & 26.88  & 53.06  & \textbf{70.06}  & 50.92  & \textbf{80.27}  & 45.59  & 41.27  & 30.03  \\
    DiffusionSat     & 76.15  & 55.22  & 26.13  & 54.64  & 69.08  & 51.32  & 79.24  & 41.76  & 43.00  & 28.72 \\
    AeroGen          & \textbf{76.94}  & 54.82  & 26.19  & 55.82  & 68.79  & 51.95  & 80.17  & 45.39  & 45.39  & 28.96 \\
    LayoutDiffusion  & 76.39  & 56.06  & \textbf{29.66}  & 55.71  & 68.70  & 51.71  & 79.13  & 47.54  & 44.93  & 29.88 \\
    MIGC             & 75.50  & 55.34  & 27.86  & 56.75  & 69.72  & \textbf{52.07}  & 79.70  & 45.11  & 46.09  & \textbf{32.99} \\
    CC-Diff (Ours)   & 76.23  & \textbf{56.13}  & 27.78  & \textbf{57.48}  & 68.09  & 51.79  & 79.37  & \textbf{49.03}  & \textbf{47.85}  & 31.85 \\

    \Xhline{1.0pt}

    \end{tabular}
}
\end{table*}

\begin{table*}[!t]
\caption{Detailed trainability results (measured by average precision)  on DOTA.}
\label{table:train_ap_dota}
\vspace{-5pt}
\centering
\resizebox{0.95\textwidth}{!}{
    \ra{1.15}
    \begin{tabular} {l ccccccc}
    \Xhline{1.0pt}
    
    Method & plane & ship & storage-tank & baseball-diamond & tennis-court & basketball-court & ground-track-field \\
    
    \hline
    Baseline         & 50.68  & 31.73  & 30.01  & 36.16  & 77.95  & 34.21  & 41.80  \\
    Txt2Img-MHN      & 50.31  & 32.64  & 30.90  & 36.16  & 77.84  & 34.32  & 44.23  \\
    DiffusionSat     & 50.63  & 32.23  & 30.92  & \textbf{36.55}  & 78.12  & 36.79  & 41.41  \\
    AeroGen & 51.15 & 33.77 & 27.20 & 37.00 & 80.80 & 37.80 & 43.57  \\
    LayoutDiffusion  & 49.67  & 31.55  & 29.67  & 36.16  & 78.22  & 34.38  & 41.40  \\
    MIGC             & 49.65  & 31.58  & 31.04  & 35.37  & 78.25  & 38.01  & 44.22  \\
    CC-Diff (Ours)   & \textbf{51.80}  & \textbf{35.39}  & \textbf{30.05}  & 38.75  & \textbf{82.46}  & \textbf{43.96}  & \textbf{45.15}  \\

    \Xhline{1.0pt}

    \end{tabular}
}
\end{table*}

\begin{table*}[!t]
\vspace{-5pt}
\centering
\resizebox{0.95\textwidth}{!}{
    \ra{1.15}
    \begin{tabular} {l cccccccc}
    \Xhline{1.0pt}
    
    Method & harbor & bridge & large-vehicle & small-vehicle & helicopter & roundabout & soccer-ball-field & swimming-pool \\
    
    \hline
    Baseline        & 38.63  & 24.13  & 26.99  & 24.64  & 25.86  & 33.64  & 38.67  & 17.94  \\
    Txt2Img-MHN     & 38.72  & 23.76  & 27.60  & 23.56  & 28.08  & 32.85  & 38.94  & 18.69  \\
    DiffusionSat    & 38.87  & 24.25  & 27.95  & 23.96  & \textbf{29.74}  & 33.99  & 39.12  & 17.67 \\
    AeroGen         & 39.23  & 23.85  & \textbf{28.87}  & 24.79  & 27.99  & \textbf{34.09}  & 40.89  & 18.83 \\
    LayoutDiffusion  & 38.38  & 25.27  & 25.90  & 24.42  & 25.78  & 31.51  & 38.10  & 16.83 \\
    MIGC             & 38.33  & 23.49  & 25.63  & 24.27  & 27.39  & 32.41  & 40.55  & 18.72 \\
    CC-Diff (Ours)   & \textbf{40.16}  & \textbf{25.09}  & 27.76  & \textbf{25.46}  & 20.08  & 31.94  & \textbf{42.62}  & \textbf{19.80} \\

    \Xhline{1.0pt}

    \end{tabular}
}
\end{table*}

\end{document}